\documentclass[]{elsarticle}

\usepackage{lineno,xcolor}
\usepackage{amsmath,amssymb}
\usepackage{siunitx} 

\usepackage{enumitem}

\usepackage{wrapfig}
\usepackage{lscape}
\usepackage{rotating}
\usepackage{epstopdf}

\usepackage{tablefootnote}
\usepackage{soul}
\usepackage{subcaption} 
\usepackage{comment}
\usepackage{multirow}
\usepackage{booktabs}
\usepackage{comment}
\usepackage{xcolor}

\usepackage[pagebackref=true,breaklinks=true,letterpaper=true,colorlinks,bookmarks=false]{hyperref}

\newcommand{\etal}{\textit{et al.}}
\newcommand{\eg}{\textit{e.g.,}}
\newcommand{\ie}{\textit{i.e.,}}

\newcommand{\xdownarrow}[1]{%
  {\left\downarrow\vbox to #1{}\right.\kern-\nulldelimiterspace}
}

\begin{document}

\begin{frontmatter}

\title{Generating Evidential BEV Maps in Continuous Driving Space}

\author[IKG]{Yunshuang Yuan}

\author[ITC]{Hao Cheng}

\author[ITC]{Michael Ying Yang}

\author[IKG]{Monika Sester}

\address[IKG]{Institute of Cartography
and Geoinformatics, Leibniz University Hannover, Germany}

\address[ITC]{Scene Understanding Group, ITC Faculty, University of Twente, The Netherlands}

\begin{abstract}
    Safety is critical for autonomous driving, and 
    one aspect of improving safety is to accurately capture the uncertainties of the perception system, especially knowing the unknown. 
    Different from only providing deterministic or probabilistic results, \eg~probabilistic object detection, that only provide partial information for the perception scenario, we propose a complete probabilistic model named GevBEV. It interprets the 2D driving space as a probabilistic Bird's Eye View (BEV) map with point-based spatial Gaussian distributions, from which one can draw evidence as the parameters for the categorical Dirichlet distribution of any new sample point in the continuous driving space. The experimental results show that GevBEV not only provides more reliable uncertainty quantification but also outperforms the previous works on the benchmarks OPV2V and V2V4Real of BEV map interpretation for cooperative perception in simulated and real-world driving scenarios, respectively. A critical factor in cooperative perception is the data transmission size through the communication channels. GevBEV helps reduce communication overhead by selecting only the most important information to share from the learned uncertainty, reducing the average information communicated by 87\%  with only a slight performance drop. Our code is published at \url{https://github.com/YuanYunshuang/GevBEV}.
\end{abstract}

\begin{keyword}
\texttt{ 
Evidential deep learning \sep 
Semantic segmentation\sep 
Cooperative perception\sep
Bird's eye view
}
\end{keyword}

\end{frontmatter}

\section{Introduction}\label{sec:intro}
In recent decades, a plethora of algorithms, \eg~\cite{lin2021local,FengHWD22,zhang2023perception,fang2022joint,zang2017lane}, have been developed for the perception systems of autonomous vehicles (AV) and many photogrammetry and remote sensing tasks. Thanks to the open-sourced datasets, \eg~\cite{kitti, nuscenes, waymo}, and the corresponding benchmarks for different standardized perception tasks for interpreting the collected data, \eg~object detection and semantic segmentation, we are able to evaluate the performance of these algorithms by comparing their predicted results with human-annotated ground truth. However, is an algorithm's better performance on these benchmarks the only goal we should chase? Obviously not; this goal is not enough for the system to be deployed reliably in the real world. 

An AV system must accurately evaluate the trustworthiness of its interpretation of the driving environment, not just focus on accuracy compared to ground truth.
This is because real-world driving is complex, with inevitable noise, limitations of sensors and algorithms, and occlusions that make it impossible for an AV system to be perfect.
For example, from an AV's perception range, occlusions are inevitable.
In this case, predicting an occluded area as a drivable road surface is likely to improve the overall accuracy because the road surface is more frequently seen in the data collected in the past, whereas it does not necessarily reflect the real-world situation.
The unaccountable guess over an unobserved target based on the prior distribution drawn from historical data ignores observation uncertainties and may even lead to serious accidents.
Consequently, the perception algorithms may evolve to reach higher scores on these benchmarks at the sacrifice of being overconfident and predicting dangerous false positives.
Therefore, safer and more trustworthy interpretations of the driving scenarios are merited.

Achieving safe driving while considering uncertainties in the perception system is a challenging task.
In the real world, uncertainties come from different sources \cite{Survey_unc}, such as sensor noise, imperfections, and perception model failures. When measurements are insufficient due to limited field-of-view (FoV), occlusions, or low sensor accuracy and resolution, high uncertainty is common.
One common circumstance is that an ego AV with limited FoV cannot reliably perceive the driving area, leading to a critical situation. 
In such a case, the ego AV should slow down and wait for the clearance of this uncertain area. An alternative is to exploit cooperative perception, \ie~the information seen by other road users with a different part of the view.  

This paper proposes a cooperative perception method with the consideration of uncertainty to address the limited FoV and safety problem.
The concept of cooperative perception (co-perception) is to share and fuse the information from the so-called Collective Perception Messages (CPMs) among AVs to enable seeing the areas beyond the ego vehicles' own views via Vehicle-to-Vehicle (V2V) communication. The AVs with communication abilities are called Connected Autonomous Vehicles (CAVs) (see an example in Figure \ref{fig:cpm_scenario}). 
One bottleneck of the co-perception technique to be realized in the real world is the communication overhead and time delay for real-time communication. Hence, sharing a large amount of data, \eg~raw sensory data, among the CAVs is not an optimal solution.   
Although recent works \cite{fpvrcnn, xu2022bridging, Cui2022CoopernautED,xu2022opencood} based on sharing deep features learned by neural networks have proven that co-perception can significantly improve the performance of the perception system, the CAVs should only share the most important information needed by the ego CAV to reduce the communication workload. The reason is that the congested network drops messages and then leads to a significant performance drop in the co-perception system.

This paper proposes to interpret the driving space for co-perception scenarios by BEV maps with uncertainty quantification.
Each 2D point on the driving surface in a BEV map is classified into one of the predefined categories.
Compared to object detection and semantic segmentation, this BEV interpretation gives a more comprehensive and denser overview of the driving surroundings to assist the AVs in safer driving plans. 
In this paper, we use Evidential Deep Learning (EDL) \cite{Sensoy2018EvidentialDL} to quantify the uncertainty of the classification.
Essentially, our proposed BEV map also provides quantified uncertainty values for the point-wise classification results.


More specifically, we interpret the driving scenarios with learnable point-based spatial Gaussian distributions in a continuous driving space instead of discrete grids so that any new sampled points in the observed area can draw densities from these Gaussian distributions. Each Gaussian describes the likelihood of the neighboring points belonging to the same class as the distribution center point.
The classification distribution of these newly sampled points is assumed to be a Dirichlet distribution to describe the probability and the uncertainty that the point belongs to one specific class. The sampled densities from the Gaussian distributions are then regarded as evidence of the Dirichlet distribution.
Hence, the parameters of these Gaussian distributions can be jointly learned by controlling the Dirichlet distributions of the new samples.
Our method provides a reliable uncertainty that is back-traceable and explainable for each of its prediction results.
For simplicity, we name our Gaussian Evidential BEV approach \textbf{GevBEV}.

Furthermore, the learned evidential BEV maps provide a holistic interpretation of the driving environment. Namely, apart from the confident detection of the drivable surface and other vehicles from the point-wise classification results,  the self-driving system knows what it is not sure about (detections with high uncertainty) or does not know at all (unobserved areas with no measurement). 
In the co-perception network, the evidential BEV maps serve as a critical criterion to identify the exact areas where extra information is needed from other CAVs via more efficient communication for the co-perception.
Based on this criterion, the most important information shared among the CAVs is distilled by intersecting the evidential maps associated with each CAV's detection in the local frame of the ego CAV. In this way, the redundant data in CPMs is avoided to prevent the communication network from saturation and package dropping.  

In summary, the \textit{key contributions} of our proposed evidential GevBEV are:
\begin{itemize}[leftmargin=*, noitemsep]
\item We propose a Gaussian-based framework to learn holistic BEV maps in a continuous space of any resolution, which is in contrast to previous works that are limited to the map resolution provided by the training data. 
\item The evidential deep learning with Dirichlet evidence is utilized to quantify the classification uncertainty and generates better-calibrated uncertainties than conventional deterministic models. 
\item Our model GevBEV achieves a new state-of-the-art performance on the co-perception benchmarks OPV2V~\cite{xu2022opencood} with simulated driving scenes as well as V2V4Real~\cite{xu2023v2v4real} in real-world driving, outperforming the runner up model CoBEVT~\cite{xu2022cobevt} with a big margin.
\item To our best knowledge, we are the first to apply evidential BEV maps for a co-perception task. Classification uncertainties serve as a critical criterion to effectively select and share CPM among CAVs and significantly reduce communication overhead. 
\end{itemize}

\section{Related Work}
\label{sec:relatedwork}

In this section, we discuss the related work in three aspects: interpretation of driving spaces, the state-of-the-art of co-perception, and uncertainty estimation of BEV maps.

\subsection{Interpretation of driving spaces}
\label{sub:Interpretations}
Object detection~\cite{overview_Jiao2019OD,li2022domain} is a typical way to interpret the driving space of an AV.
A detected object is commonly characterized by a 2D/3D bounding box from camera/LiDAR data.
However, this interpretation of the driving scenario may not be complete because the space with no detection or occlusion is not interpreted.
This space can be some drivable areas, non-drivable areas, or be occupied by objects that are not detected or observed.
Consequently, the AV may not be able to make reliable driving decisions depending on the output of object detection. 

Semantic segmentation is another common method for the interpretation of the driving space. 
It classifies each measurement point -- pixels in images or points of LiDAR reflections -- into a specific semantic class. 
To further differentiate the points from the same semantic class but belonging to different object instances, it is extended to panoptic segmentation~\cite{Kirillov2018PanopticS}; besides the semantic label, the measurement point is also assigned with an instance identity. 
Although semantic and panoptic segmentations are holistic and dense in the ego AV's perspective -- range view, they are partial and sparse in the BEV, from which the AV usually makes driving plans \cite{qiu2022gfnet}. 

Typically, the interpretation of the driving space is further extended to BEV map segmentation to mitigate the aforementioned limitations.
The driving environment is represented as a BEV 2D image \cite{cvt_zhou2022, xu2022cobevt} and each pixel in the BEV map is marked with a semantic label, which gives a holistic overview of the driving surface for vehicle mapping and planning \cite{bev_Loukkal2021Flatmobiles}. 
In previous image-based works, BEV interpretations are also carried out as occupancy grid mapping \cite{bev_Lu2019MonocularSO}, cross-view semantic segmentation \cite{bev_Pan2020CrossViewSS}, or map-view semantic segmentation \cite{cvt_zhou2022}. 
They transform image features from the image coordinates to an orthographic coordinate of the BEV map via either explicit geometric or implicit learned transformations \cite{cvt_zhou2022,xu2022cobevt,li2022bevformer}. Compared to images, point cloud data with 3D information are more straightforward to generate such BEV maps by compressing information in the orthogonal direction in approaches such as PIXOR \cite{Yang2018PIXORR3}, Pointpillars \cite{Lang2019PointPillarsFE}, and Voxelnet \cite{Zhou2018VoxelNetEL}. We also resort to BEV maps for a holistic view of the 2D driving space.
Nevertheless, because of the sparsity of distant measurements and occlusions, generating dense BEV maps from images or point clouds is unreliable without considering observability. For example, the occupied area of some occluded vehicles might be classified as a drivable area just because the categorical distribution learned from the historical data implies that the invisible points in the BEV map are more likely to be a drivable area than a vehicle.
Therefore, in order to avoid unaccountable predictions on unobserved areas, we propose to only draw results from observed areas based on the geometric location of the measured points.

\subsection{State-of-the-art of  co-perception}

With the development of Vehicle-to-Vehicle communication and the availability of simulation tools to generate high-fidelity collaborative detection data \cite{dosovitskiy2017carla,xu2021opencda,xu2022opencood}, co-perception extends the perception system from a single ego vehicle's perspective to including the perceptions from neighboring vehicles. Modern deep learning architectures, such as graph neural networks and Transformer~\cite{vaswani2017attention}, are utilized to fuse the perception information, mainly deep features of the backbone detection networks, from the CAVs. For example, V2Vnet~\cite{wang2020v2vnet} and DiscoNet~\cite{li2021learning} use graph models to aggregate the detection information from nearby vehicles. AttFuse~\cite{xu2022opencood}, V2XViT~\cite{xu2022v2x}, and CoBEVT~\cite{xu2022cobevt} propose to use the Transformer network with the self-attention mechanism to facilitate the collaboration and information fusion in a BEV setting among CAVs.
FCooper~\cite{Fcooper} keeps the sparsity of feature maps learned from point clouds to reduce communication overhead of co-perception, and uses Maxout to fuse these features.
However, none of these methods have explored uncertainty estimation to filter out non-informative data shared among the CAVs and further increase the communication efficiency of the co-perception system.

\subsection{Uncertainty estimation of BEV maps}
\label{subsec:uncertainty}
Uncertainties are not avoidable in DNNs, making it necessary to estimate them, especially for safety-critical applications. The uncertainty of a DNN's output is called predictive uncertainty~\cite{Survey_unc}. This uncertainty is mostly qualified either by modeling the epistemic uncertainty that captures the systematic uncertainty in the model or the aleatoric uncertainty that captures the random noise of observations \cite{Kendall2017WhatUD}. There are also other approaches, such as Prior Network \cite{priornet_Malinin2018} that quantifies the predictive uncertainty by modeling the distributional uncertainty caused by the distribution mismatch between the training data and the new inference data.

To estimate the epistemic uncertainty, Bayesian Neural Networks (BNNs) \cite{bnn1_Mackay1991,bnn2_Neal1995} provide a natural interpretation of the uncertainty by directly inferring distributions over the network parameters. Notwithstanding, they are hard to use for DNNs because calculating the posterior over millions of parameters is intractable. Therefore, approximation methods such as Monte-Carlo (MC) Dropout \cite{Gal2016Unc} and Deep Ensemble \cite{DeepEnsem} have been developed. MC Dropout shows that training a dropout-based neural network is equivalent to optimizing the posterior distribution of the network output. However, several forward runs have to be conducted with the dropout enabled to infer the uncertainty, which is inefficient and time-consuming.
Therefore, it is not considered in this work. Deep Ensemble trains several models to approximate the distribution of the network parameters. It also needs several forward runs over each trained model and thus is also not adopted in this work for the same reason.

To capture the aleatoric uncertainty, Direct Modeling is widely used \cite{DM_Feng2019onLidar,DM_Meyer2019LaserNet,DM_Miller2019EvaluatingMS,DM_Pan2020TowardsBP,MD_Feng2020CamLidar}. Compared to MC Dropout and Deep Ensemble, Direct Modeling assumes a probability distribution over the network outputs and directly predicts the parameters for the assumed distribution. Therefore, uncertainty is obtained over a single forward run and is more efficient. For classification problems, the conventional deterministic DNNs apply the Softmax function over the output logits to model the categorical distribution as multi-nominal distribution. However, the Softmax outputs are often overconfident and poorly calibrated \cite{Sensoy2018EvidentialDL,cls_unc2019TowardsBC}.

Instead, converting the output logits into positive numbers via, \eg~ReLU activation to parameterize a Dirichlet distribution quantifies class probabilities and uncertainties better.
For example, the Prior Network \cite{priornet_Malinin2018} captures the predictive uncertainty by explicitly modeling the distributional uncertainty and minimizing the expected Kullback-Leibler (KL) divergence between the predictions over certain (in-distribution) data and a sharp Dirichlet and between the predictions over uncertain (out-of-distribution) data and a flat Dirichlet. However, additional out-of-distribution samples are needed to train such a network to differentiate in- and out-of-distribution samples.
In complex visual problems like object detection and semantic segmentation, obtaining enough samples to cover the infinite out-of-distribution space is prohibitive. 

Differently, the Evidential Neural Network \cite{Sensoy2018EvidentialDL} treats the network output as beliefs following the Evidence and Dempster-Shafer theory \cite{Dempster2008AGO} and then derives the parameters for the Dirichlet distribution to model the epistemic uncertainty. 
Compared to BNNs, this method quantifies the uncertainty of a classification by the collection of evidence leading to the prediction result, meaning that the epistemic uncertainty of the classification can be easily quantified by the amount of evidence.
Instead of minimizing the discrepancy of the predictive distributions with pre-defined ground truth distributions, Evidential Neural Network formulates the loss as the expected value of the basic loss, \eg~cross-entropy, for the Dirichlet distribution. Therefore, no additional data or ground truth distributions are needed. Hence, we propose to apply this method for modeling the categorical distributions of the points in a 2D driving space. 

Moreover, instead of modeling the uncertainty with Dirichlet distribution directly, we introduce a spatial Gaussian distribution for the measurement points and draw Dirichlet distributions for a BEV map based on the predicted Gaussian parameters. This configuration mimics the conditional random field algorithm \cite{crf}, which can be used to smooth the segmentation results by considering the neighboring results.
It should be noted that in this paper, we are more focused on epistemic uncertainty (a lack of knowledge in the neural network-based model) to help us understand the output of the perception system in CAVs, and to use this uncertainty quantification in the co-perception step for distilling the most important information shared among the CAVs.

\section{Method}
\label{sec:method}
\subsection{Problem Formulation}
\label{subsec:problemform}
We formulate the task of generating the Gaussian evidence BEV (GevBEV) map with object detection and semantic segmentation under the setting of co-perception between the ego CAV and cooperative CAVs.
We follow the OPV2V benchmark~\cite{xu2022opencood} setting for co-perception.
Given the ego vehicle, there are $N_\text{coop} < 6$ cooperative CAVs in the communication range of the ego vehicle and some other vehicular participants. All CAVs can send CPMs to each other in a Request-Respond manner -- the ego CAV first sends a CPM request that specifies the information over the area it needs, then a cooperative CAV that receives this request will respond with the corresponding message only if the request information is available.

In this paper, we choose to use point cloud data to demonstrate our proposed interpretation of driving spaces. It should be noted that this interpretation can be applied to any modality or multi-modalities as far as the measurement points can be projected to the BEV map, not only for autonomous driving but also other mapping tasks with multiple sources of measurements. 
We leave this for our future work.
The input feature vector of each point in the point cloud is denoted as $f^\text{in}=[x, y, z, d, \cos{\theta}, \sin{\theta}, \mathfrak{i}]$, where $x, y, z$ are the local coordinates in the ego LiDAR frame, $d$ is the distance of the point to the LiDAR origin, $\theta$ is the angle of the point relative to the $x$-axis of the LiDAR frame, and $\mathfrak{i}$ is the intensity of the LiDAR reflection. 
This input feature vector is leveraged to train a U-Net-based~\cite{ronneberger2015u} end-to-end multi-task network. 
Namely, the main outputs of the proposed GevBEV are object detection results, the BEV maps for both the driving surface (\ie~roads) and dynamic road objects (\ie~vehicles), and the uncertainty of the predicted BEV maps is quantified by a categorical Dirichlet distribution.

In the following subsections, we first introduce the detailed framework for generating GevBEV maps by sharing all the perception information among CAVs without any filtering strategy to distill the most important information. Then, we present the method for an uncertainty-based CPM selection by manipulating our generated evidence BEV maps to improve the efficiency of the communication among CAVs.

\subsection{Framework}

\begin{figure}[t]
  \centering
    \includegraphics[width=0.8\linewidth]{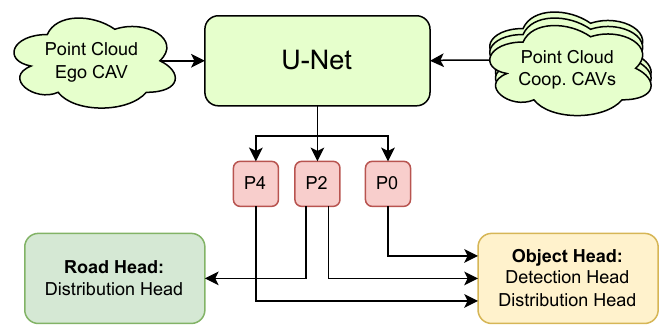}
    \caption{Overview of the GevBEV map framework. It takes as input the point cloud data. The road head learns the distribution for the road surface (green), and the objects head (orange) detects the objects and learns the corresponding distribution of the bounding boxes. 
    $P_s$: intermediate learned features of voxels downsampled with strides $s=0, 2, 4$.}
    \label{fig:pipeline}
\end{figure}

\begin{figure*}
  \centering
    \includegraphics[width=0.9\linewidth]{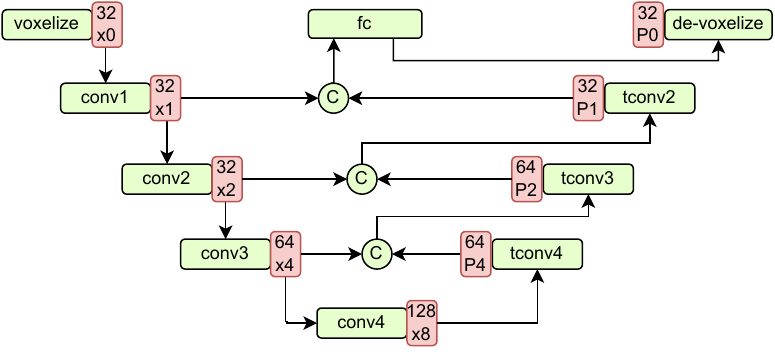}
    \caption{The U-Net-based backbone. The blue boxes indicate the network components, the orange boxes show the components' output channels and voxel strides. The green ellipses are the concatenation operation.}
    \label{fig:U-Net}
\end{figure*}

The overview of the GevBEV map framework is shown in Figure \ref{fig:pipeline}.
It takes as input the point cloud data from ego and cooperative vehicles and trains multiple heads for multi-task learning.
Concretely, a U-Net-based backbone network first learns and aggregates deep features of different resolutions from the input point clouds. The aggregated features for voxels of different sizes are notated as $P_s$, where $s$ is the downsampling ratio relative to the predefined input voxel size. $P_0$ are the learned features for each point in the point cloud.
Those features of different voxel sizes are separately input to a road head and an objects head.
The road head, specified as \textit{Distribution Head}, generates the GevBEV map layer for the static road surface; The objects head, specified as \textit{Detection Head} and \textit{Distribution Head}, detects dynamic objects and also generates the GevBEV map for the object layer. 
We separate the GevBEV map into two layers because objects tend to have smaller sizes compared to the road surface. Otherwise, small objects could be smoothed out by the points of surfaces because they are dominating in quantity.
We explain each module of the framework, \ie~\textit{U-Net-based backbone}, \textit{Detection Head}, and \textit{Distribution Head}, in the following in detail.

\paragraph{\textbf{U-Net}}
Given its high performance on point-wise feature extraction and representation, U-Net~\cite{ronneberger2015u} is utilized as the backbone of our model.
Figure \ref{fig:pipeline} depicts the general structure of the U-Net with our customized encoding. 
First, the input features are voxelized with a Multi-Layer Perceptron ($\operatorname{mlp}$) as described by Eq~\eqref{eq:voxelize1}, 
\begin{equation}\label{eq:voxelize1}
    \Bar{f}_v = \frac{\sum_{i \in \mathbf{v}} f_{i}^\text{in}}{|\mathbf{v}|},   \quad
    \Tilde{f}_{vi} = \operatorname{mlp}([f_{vi}, (f_{vi} - \Bar{f}_v)]),
\end{equation} 
where $\mathbf{v}$ is the set of points belonging to a voxel, $f_{i}^\text{in}$ denotes the input feature of point $i$, and $[\cdot, \cdot]$ represents the concatenation operation.
Then, the voxel features $\Tilde{f}_{\mathbf{v}}$ are calculated by averaging the encoded point features $\Tilde{f}_{vi}$ that belong to the voxel, as denoted in Eq~\eqref{eq:voxelize2}. 
\begin{equation}\label{eq:voxelize2}
    \Tilde{f}_{\mathbf{v}} = \frac{\sum_{i \in \mathbf{v}} \Tilde{f}_{vi}}{|\mathbf{v}|}. 
\end{equation}
Afterwards, the encoded voxels are fed to the four convolutional blocks. 
Namely, $\operatorname{conv1}$ contains only one layer to digest the input. 
$\operatorname{conv2}$ to $\operatorname{conv4}$ consist of three convolutional layers, in which a previous layer down-samples the sparse tensors.
Each sparse convolutional layer is followed by batch normalization and Leaky ReLU activation.  
In the upsampling layers, the transposed convolutional layers have a similar structure as the counterparts in the downsampling layers. 
The features from the shortcuts of the convolutional layers are all concatenated with the features from the transposed convolutional layers. 
In the end, we concatenate the voxel features with the encoded point features $\Tilde{f}_{vi}$ to de-voxelize the stride-one voxels and obtain features $P_0$ for each point.
It is worth mentioning that all the convolutional layers in this network are implemented with Minkowski Engine \cite{mink_choy20194d} over sparse voxels to decrease the computational load.

\begin{figure}[ht!]
\setlength{\tabcolsep}{2pt}
\begin{minipage}{.485\linewidth} 
    \centering
        \includegraphics[width=0.8\linewidth]{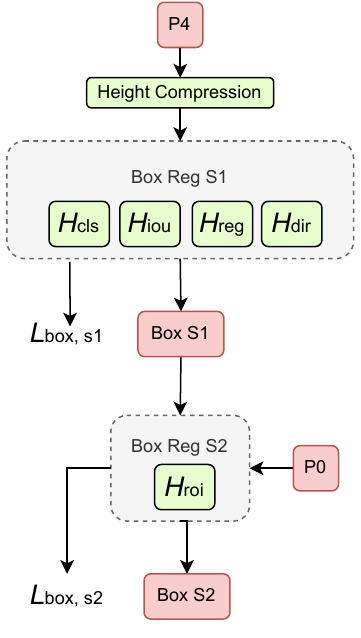}
        
        \caption{Detection head.}
        \vspace{-12pt}
        \label{fig:detection_head}    
\end{minipage} 
\hspace{6pt}
\begin{minipage}{.485\linewidth}
    \centering
        \includegraphics[width=0.8\linewidth]{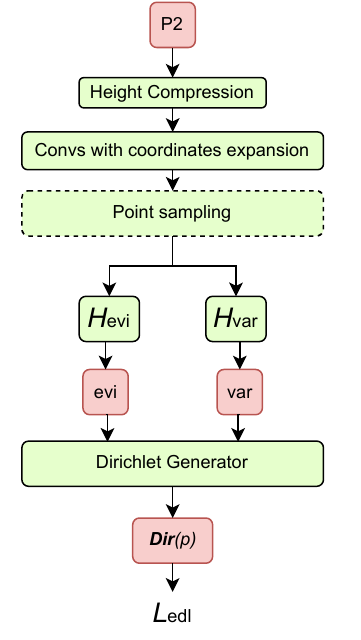}
        \caption{Distribution head.}
        \vspace{-12pt}
        \label{fig:distribution_head}
\end{minipage} 
\end{figure}
\paragraph{\textbf{Detection Head}} We use this head to demonstrate the alignment of the predicted bounding boxes of object detection with the GevBEV maps.
As shown in Figure~\ref{fig:detection_head}, in this head, we use voxels of stride-four $P_4$ to generate pre-defined reference bounding boxes, the so-called anchors, and to do further regression based on them. 
First, we use strided convolutional layers over the vertical dimension to compress the height of the 3D sparse tensor in order to obtain a BEV feature map. 
Then, this map is used for the bounding box classification and regression. 
We follow \cite{fpvrcnn} that uses a two-stage strategy for co-perception: Stage-one (BoxReg S1) generates proposal boxes based on the local information of the ego CAVs, and Stage-two (BoxReg S2) refines the boxes by fusing the information from the neighboring CAVs.
  
We use the following Eq.~\eqref{eq:box_enc1}-\eqref{eq:box_enc3} to encode bounding boxes:
\begin{equation}\label{eq:box_enc1}
    \mathcal{B}_\textit{loc} = [\frac{x_g-x_a}{d_{xy}}, \frac{y_g-y_a}{d_{xy}}, \frac{z_g-z_a}{h_a}],
\end{equation}
\begin{equation}\label{eq:box_enc2}  
    \mathcal{B}_\textit{dim} = [\log (\frac{l_g}{l_a}), \log (\frac{w_g}{w_a}), \log (\frac{h_g}{h_a})],
\end{equation}
\begin{equation}\label{eq:box_enc3} 
\begin{aligned}
    \mathcal{B}_\textit{dir} &= [\cos{\theta_g} -  \cos{\theta_a}, \sin{\theta_g} - \sin{\theta_a},\\
    &\cos{\theta_g} - \cos{(\theta_a + \pi)}, \sin{\theta_g} - \sin{(\theta_a + \pi)}].    
\end{aligned}
\end{equation}
$\mathcal{B}_\textit{loc}$ and $\mathcal{B}_\textit{dim}$ are the commonly used encodings for the location and dimension of the bounding boxes. The subscripts $g$ and $a$ represent ground truth and anchor, respectively. $x,y,z$ are the coordinates of the bounding box centers, and $d_{xy}$ is the diagonal length of the bounding box in the $XY$-plane. $l, w, h$ indicate the bounding box length, width, and height. 
We use $\sin$ and $\cos$ encodings, denoted by Eq~\eqref{eq:box_enc3}, for the bounding box direction $\theta$ to avoid the direction flipping problem. The first two elements of $\mathcal{B}_\textit{dir}$ are the encodings of the original direction angle, while the last two elements are for the reversed direction. 
The additional encodings for the reversed direction reduce the distance of sin-cos encodings between the regression source--anchors and the target--ground truth boxes whenever they have opposite directions, therefore, facilitating the direction regression.
 
The overall loss for all the heads in BoxReg S1 is notated as $\mathcal{L}_\textit{box, s1}$, as shown in Figure \ref{fig:detection_head}. To be more specific, BoxReg S1 follows CIASSD \cite{ciassd} to use four heads $\mathcal{H}_\textit{cls}, \mathcal{H}_\textit{iou}, \mathcal{H}_\textit{reg}, \mathcal{H}_\textit{dir}$ for generating bounding boxes. $\mathcal{H}_\textit{cls}$ uses binary cross-entropy, other heads in this stage use smooth-L1 loss. $\mathcal{H}_\textit{iou}$ regresses the Intersection over Unions (IoU) between the detected and ground truth bounding boxes. It is used to rectify the classification score from $\mathcal{H}_\textit{cls}$ so that better-bounding boxes can be kept during Non-Maximum Suppression (NMS). The direction head $\mathcal{H}_\textit{dir}$ regresses the encoded angle offsets between the ground truth bounding boxes and the corresponding anchors. During the box decoding phase,  the predicted angle offsets from $ \mathcal{H}_\textit{dir}$ with smaller values are selected as the correct direction of the bounding boxes.

In BoxReg S2, the proposal bounding boxes detected by different CAVs are first shared with the ego CAV and fused by NMS, and then they are used as reference anchors for further refinement. 
Concretely, following FPV-RCNN \cite{fpvrcnn}, the CAVs only share the keypoints belonging to the proposals and then use these keypoints to refine the fused bounding boxes. 
Moreover, in this paper, we use Minkowski Engine \cite{mink_choy20194d} to simplify the structure of Region-of-Interest (RoI) head $\mathcal{H}_\textit{roi}$. 
Namely, the fused boxes are first transformed to a canonical local box coordinate system, where the keypoints coordinates are noted as $\text{kpt}^{in}_i$. 
We then voxelize the keypoints to a $6\times6\times6$ grid with a similar voxelization operation as used in the U-Net, where $f^\text{in}_i$ becomes the canonical coordinates $\text{kpt}^{in}_i$ of keypoints $[x, y, z]$. 
The first element $f_{vi}$ of the concatenation in Eq~\eqref{eq:voxelize1} becomes the learned feature of keypoints and the second element becomes the learned positional encoding of the keypoint coordinates. 
To make it straightforward for understanding, Eq~\eqref{eq:voxelize1} is reformulated into Eq~\eqref{eq:voxelize3}. 
\begin{equation}\label{eq:voxelize3}
\begin{aligned}
    \Bar{\text{kpt}}_v &= \frac{\sum_{i \in \mathbf{v}} \text{kpt}_{i}^{in}}{|\mathbf{v}|},   \\
    \Tilde{f}_{vi} &= \operatorname{mlp}([f_{vi}, \operatorname{mlp}(\text{kpt}_{vi} - \Bar{\text{kpt}}_v)]).    
\end{aligned}
\end{equation} 
After summarizing the features for each voxel by Eq~\eqref{eq:voxelize2}, all voxels in the grids are then aggregated by the weighted average of the voxel features $\Tilde{f}_\textbf{v}$. 
The weights are learned from $\Tilde{f}_\textbf{v}$ by $\operatorname{mlp}$. 
The aggregated features of each box grid are then fed to one IoU head and one box regression head to generate the refinement parameters for the fused proposal boxes. 
The decoded bounding boxes are notated as Box S2, as shown in Figure \ref{fig:detection_head}.
Smooth-L1 loss is used for both heads. The summation of the losses in this stage is notated as $L_\textit{box, s2}$.

\paragraph{\textbf{Distribution Head}} 
As illustrated in Figure \ref{fig:pipeline}, we design a distribution head to generate point-based distributions for both road surface and objects. To balance the trade-off between computational load and the point resolution, we use voxels of stride-two $P_2$ to generate distributions for this purpose. The overall structure of the distribution head is described in Figure \ref{fig:distribution_head}. 
The input 3D voxels are first compressed along the vertical direction to obtain the 2D point-wise deep features. This height compression is composed of two sparse convolution layers with the same kernel size and stride size so that all voxels in the vertical direction can merge into one. In this module, the weights are shared for both the road and object distribution heads.

Furthermore, we dilate the voxel coordinates during the sparse convolutions to close the gaps between discrete measurement points. We call this as coordinate expansion.
Even though we propose to only predict the distributions on observable areas, all sensors measure the continuous space in a discrete way, leading to unavoidable gaps at a large measure distance.
As described above,  on the one hand, the basic convolutional layers used in the U-Net maintain the sparsity of the voxels to reduce the computational load. On the other hand, they cannot infer information from the gaps between two neighboring laser measurements caused by the range view of the LiDAR at distant observable areas.  
However, as long as the measurement density is enough, the model should also be able to infer the information between the two discrete measurements in a controlled manner. Therefore, we use several coordinate-expandable sparse convolutions to carefully close the gaps by controlling the expansion range with predefined kernel sizes. The detailed setting is given in Sec.~\ref{subsubsec:implementation}. 

To further accelerate the training process, we use the point sampling module, as illustrated with the dashed line box in Figure \ref{fig:distribution_head}, to downsample the total amount of points.
We call all these selected points center points for the simplicity of explanation.
This module is optional, and the later empirical results show that it does not have an obvious negative influence on the training result.  

We assume each center point $c_i$ has a Dirichlet distribution to model the point classification distribution and a spatial isotropic Gaussian distribution to model the neighborhood consistency of the point. 
The parameters for these two distributions are then regressed by the head $\mathcal{H}_\text{cls}$ and $\mathcal{H}_\text{var}$, respectively. Both heads are composed of two fully connected layers activated by ReLU to constrain the parameters for both distributions to be positive. Their outputs are noted as
\begin{align}
    \mathbf{o}_\text{cls} &= [o^\text{fg}_\text{cls}, o^\text{bg}_\text{cls}], \\
    \mathbf{o}_\text{var} &= [o^\text{fg}_{\sigma x}, o^\text{fg}_{\sigma y}, o^\text{bg}_{\sigma x}, o^\text{bg}_{\sigma y}],
\end{align}
where $\text{fg}$ indicates foreground and $\text{bg}$ background. $\mathbf{o}_\text{cls}$ is regarded as the evidence of the point to be foreground or background. $\mathbf{o}_\text{var}$ is the regressed variances of the point in $x$- and $y$-axis. To ensure that each point is contributing, we add a small initial variance to the predictions. Hence, the resulting variances $\mathbf{\sigma}^2_{x,y}=\mathbf{o}_\text{var}+\mathbf{\sigma}_0^2$. 
For any new given target point $\mathbf{x}_{j}$ in the neighborhood of the center point $\mathbf{c}_i$ in the BEV space, we can then draw the probability density $\phi(\mathbf{x}_{j})_i$ of this new point belonging to a specific class by Eq~\eqref{eq:prob}, 
\begin{align}
    \Sigma_i &= 
    \begin{bmatrix}
    \sigma_x^2, 0\\
    0, \sigma_y^2
    \end{bmatrix},\\
    m_{ji} &= (\mathbf{x}_j - \mathbf{c}_i)^T\Sigma^{-1}_i(\mathbf{x}_j-\mathbf{c}_i), \\
    \phi(\mathbf{x}_{j})_i &= \frac{\exp{(-0.5\cdot m)}}{\sqrt{2\pi^d |\Sigma_i|}},\label{eq:prob}
\end{align}
where $\Sigma_i$ is the covariance of the center point $\mathbf{c}_i$ for foreground or background distribution, $m$ is the squared Mahalanobis distance of point $\mathbf{x}_j$ to the center point $\mathbf{c}_i$, and $d$ is the dimension of the distribution. In our case, $d=2$.

To obtain the overall Dirichlet evidence $e(\mathbf{x}_j)$ for point $\mathbf{x}_j$, we summarize the normalized and weighted probability mass drawn from all the neighboring center points $\operatorname{nbr}(j)$ that is in the maximum distribution range $\nu$ as  
\begin{equation} \label{eq:evi}
\begin{aligned}
    e_k(\mathbf{x}_j) &= \sum_{i \in \operatorname{nbr}(j)} \frac{\phi(\mathbf{x}_{j})_i}{\phi(\mathbf{c}_i)} \cdot o^\textit{k}_{cls}, \\
    &= -\frac{1}{2} \sum_{i \in \operatorname{nbr}(j)} m_{ji} \cdot o^\textit{k}_\text{cls},
\end{aligned}
\end{equation}
where $\phi(\mathbf{c}_i)$ is the probability density at the center point, and $k \in \{\text{fg}, \text{bg}\}$. 
Hereafter, Eq~\eqref{eq:evi} is derived as following
\begin{equation} \label{eq:prob_div}
\begin{aligned}
    \log (\phi(\mathbf{x}_{j})_i) &= -\frac{1}{2}m_{ji} - \frac{1}{2} \log (2\pi^d |\Sigma|), \\
    &= -\frac{1}{2}(d\log(2\pi) - \log|\Sigma|) - \frac{1}{2}m_{ji}, \\
    \log (\phi(\mathbf{c}_{i})) &= -\frac{1}{2}m_{i} - \frac{1}{2} \log (2\pi^d |\Sigma|), \\
    &= -\frac{1}{2}(d\log(2\pi) - \log|\Sigma|), \\
    \frac{\phi(\mathbf{x}_{j})_i}{\phi(\mathbf{c}_i)} 
    &= \exp (\log \frac{\phi(\mathbf{x}_{j})_i}{\phi(\mathbf{c}_i)}), \\
    &= \exp (\log \phi(\mathbf{x}_{j})_i -  \log \phi(\mathbf{c}_{i})), \\
    &= -\frac{1}{2}m_{ji},
\end{aligned}
\end{equation}
where the squared Mahalanobis distance of the center point $\textbf{c}_i$ to itself is $m_{i}=0$. Following \cite{Sensoy2018EvidentialDL}, the expected probability $p_{j,k}$ -- point $\mathbf{x}_{j}$ belonging to class k -- and the uncertainty $u_j$ of this classification result are
\begin{equation}\label{eq:unc}
    \hat{p}_{j,k} = \frac{\alpha_{j,k}}{S_j}
    = \frac{e_k(\mathbf{x}_j) + 1}{\sum _{k\in\{\text{fg},\text{bg}\}}(e_k(\mathbf{x}_j) + 1)},
\end{equation}
\begin{equation}
    u_{j} = \frac{K}{S_j},
\end{equation}
where $\alpha_{j,k}$ is the concentration parameter of class $k$ for $k = 1, ..., K$, and $S_j$ the strength of the Dirichlet distribution of point $\mathbf{x}_j$. For the loss $\mathcal{L}_\textit{edl}$ of the distribution head, we use the recommended loss function in \cite{Sensoy2018EvidentialDL}, which is formulated as the expectation of the sum of the squared loss and a Kullback-Leibler (KL) divergence regularization that prevents the network from generating excessively high evidences. 
The final expression of this loss is described by 
\begin{equation}\label{eq:edl}
\begin{aligned}
    \mathcal{L}_{\textit{edl}} &= \sum_{j=1}^N \sum_{k\in\{\text{fg},\text{bg}\}} [(y_{j,k} - \hat{p}_{j,k})^2 + \frac{\hat{p}_{j,k}(1-\hat{p}_{j,k})}{S_j+1} ] \\
    &+ \lambda_t \sum_{j=1}^N \operatorname{KL}[\operatorname{Dir}(\mathbf{p}_j|\Tilde{\mathbf{\alpha}_j}) \parallel \operatorname{Dir}(\mathbf{p}_j|\mathbf{1})],
\end{aligned}
\end{equation}
where
\begin{equation*}
\begin{aligned}
     \Tilde{\mathbf{\alpha}_j} &= \mathbf{\alpha}_{j} \odot (1-\mathbf{y}_j) + \mathbf{y}_j,\\
    \lambda_t &= \min(1, A_\text{epoch} / A_\text{max}).
\end{aligned}
\end{equation*}
The first term in Eq~\eqref{eq:edl} is the expected sum of squared loss between the target label $y_{j,k}$ and the prediction $p_{j,k}$. $N$ denotes the total number of samples, and $j\in \{1, ..., N\}$.
The second term is the KL-divergence weighted by an annealing coefficient $\lambda_t$ that changes with the ratio between the epoch number $A_\text{epoch}$ and the maximum annealing step $A_\text{max}$. $\mathbf{y}_j=\{y_{j,k}\}_ {k\in\{\text{fg},\text{bg}\}}$ is the categorical ground truth label of point $\mathbf{x}_j$. $\Tilde{\mathbf{\alpha}_j}$ is the filtered version of $\mathbf{\alpha}_{j}$ to ensure that the KL-divergence only punishes the misleading predictions of $\mathbf{\alpha}_{j}$.

We propose a Gaussian-based method to train the Distribution Head in a continuous space for the evidential BEV map.
To achieve this goal, the target points for the supervised learning are not limited to the original observation points of the point clouds or the center points of a specific resolution with discrete points;
They can be any points in the continuous BEV plane.
More specifically, the original observed target points are treated as seeds based on which we generate continuous target points by randomly shifting them from their original observation in a controlled range, \eg~by a normally distributed distance $\mathcal{N} (0, 3)\,m$.
In this way, a pre-defined number $n_\text{tgt}$ of target points are generated from the center points to control the density, and only the generated target points of the observed areas are leveraged to train the model.
To avoid memory overflow and long-computational time during training, the randomly sampled target points are further down-sampled.
Via voxel down-sampling, the road head is supervised only with a limited number of target points -- $N_\text{tgt}$ for both foreground and background samples.
However, the importance and the amount of target points for objects head are more biased between the foreground and background. Only a small amount of background target points are included for training as they are the majority but less important.  In contrast, all samples that are extended to a specific range of the ground truth bounding boxes' edges are adopted for training the foreground class and better describing the details around the bounding boxes. 

\subsection{Co-Perception} \label{sec:co_perception}
\begin{figure}[t]
  \centering
    \includegraphics[width=0.7\linewidth]{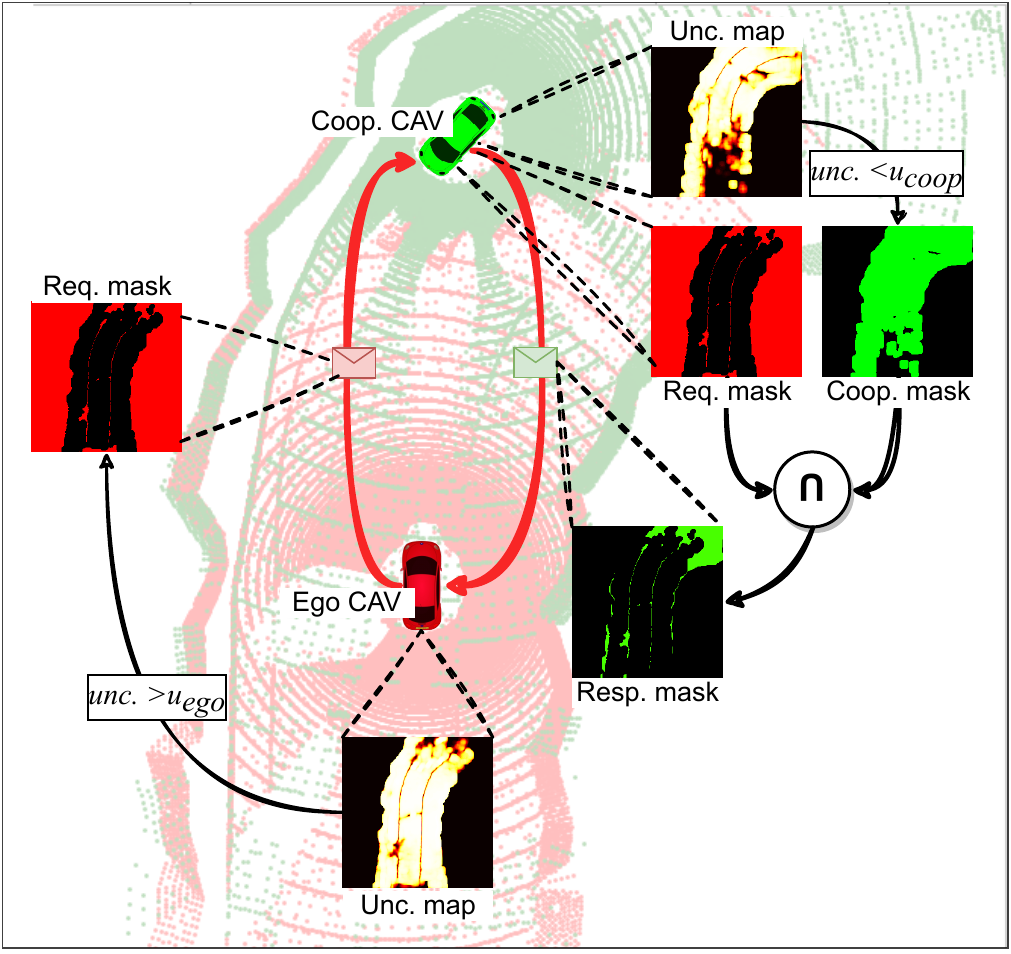}
    \caption{An example of the information shared between CAVs. The ego CAV and its request mask with high uncertainty areas indicated in red color, and the cooperative CAV and its response mask with low uncertainty indicated in green color. In the uncertainty maps, light color indicates low uncertainty and black color indicates no measurements.}
    \label{fig:cpm_scenario}
\end{figure}


This section introduces the co-perception method for the uncertainty-based CPM selection by manipulating our generated evidence BEV maps.
Instead of simply sharing all the information among CAVs that may congest the communication network, the most important information is distilled by the generated BEV maps that quantify the perception uncertainty of the CAVs.
As shown in Figure \ref{fig:cpm_scenario}, the red ego CAV first generates a request binary mask by thresholding its uncertainty map generated by Eq~\eqref{eq:evi} with the threshold $u_\text{ego}$ (bottom Unc. mask) and only sends the request for the perception information in the areas it has high uncertainty (left red Req. maks). Then, the green cooperative CAV responds with its own masked evidence map(green Resp. mask). To be more specific, the binary mask generated by thresholding the uncertainty map of the cooperative CAV with the threshold $u_\text{coop}$ is used to intersect with the received request mask from the ego CAV, resulting in a response mask with the low-uncertainty areas from the cooperative CAV.
Afterwards, this resulting response mask is used to distill the CPM communicated from the cooperative CAV to the ego CAV by only selecting the evidences over the response-masked areas.
We denote this sharing strategy as $CPM_\text{all}$ because it considers all areas in the pre-defined FoV of the ego CAV for information sharing. 
However, in the driving space, the CAVs pay more attention to the situations on the road surface. 
To this end, the information to be shared can be further constrained by the road surface geometry of the current scenario. 
In real applications, this geometry can be retrieved from some prior information, such as maps.
We notice that the co-perception benchmarks also provide an HD map acquired beforehand.
As a proof-of-concept study, we register the current scenario to the HD map to further rule out non-surface areas in the masked areas.
We denote this sharing strategy as $CPM_\text{road}$ when the extra HD map is already provided to the CAVs.
For simplicity, in this paper, we fix the uncertainty threshold $u_\text{coop}=1.0$ for the cooperative CAVs and only vary the threshold $u_\text{ego}$ for the ego CAV in our experiments to evaluate its effectiveness. 

\section{Experiments}
\label{sec:experiment}
In the following, we introduce the dataset, data augmentations, evaluation metrics, and detailed experiments to evaluate our proposed model. 

\subsection{Dataset} 
In this work, we conduct the experiments on two multi-agent co-perception benchmarks,  OPV2V~\cite{xu2022opencood}, a simulated dataset generated by CARLA~\cite{dosovitskiy2017carla} and OpenCDA~\cite{xu2021opencda,xu2023opencda}, and V2V4Real~\cite{xu2023v2v4real}, a real dataset captured with two vehicles driving in real-world scenarios.

The OPV2V dataset has 73 scenes, including six road types from nine cities. It contained 12K frames of LiDAR point clouds and the annotated 3D bounding boxes for each frame.
The detection range of this dataset is set to $[-50, 50]\,m$ for $x$- and $y$-coordinate and $[-3,3]\,m$ for $z$ coordinate, same as the the baseline work CoBEVT~\cite{xu2022cobevt}.
The V2V4Real dataset covers a driving area of \SI{410}{km}. It contains about 10K annotated LiDAR frames. We set the detection range of this dataset to $[-102.4, 102.4]\,m$ for $x$-coordinate, $[-38.4, 38.4]\,m$ for $y$-coordinate and $[-5, 3]\,m$ for $z$-coordinate as most annotated vehicles are in this range. 
We follow the official partitioning of both datasets for the training and test.
Namely, 44 training scenarios and 16 test scenarios for OPV2V, 33 training scenarios and 9 test scenarios for V2V4Real.

\subsection{Data Augmentation}
A point cloud is one of the most common ways to represent the geometric information sensed by LiDAR sensors. 
It is a collection of reflected points when the LiDAR rays hit the surface of objects.
Considering that the points are very sparse if a ray hits distant objects in a point cloud, we propose the following augmentations of the LiDAR data to help the perception tasks.

\begin{figure}[t]
  \centering
    \includegraphics[width=0.5\linewidth]{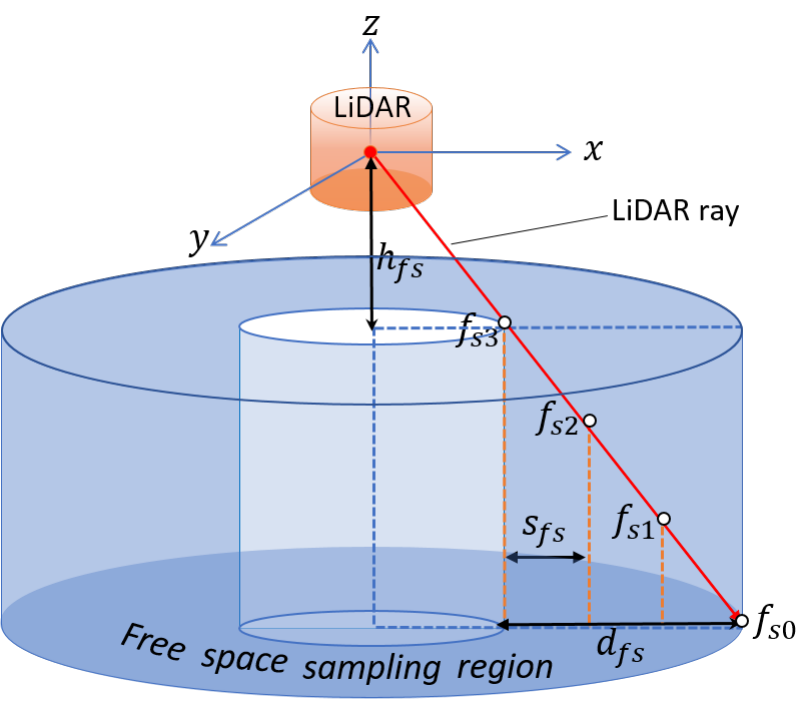}
    \caption{Sampling free space points. The red point is the origin of the LiDAR coordinate system, $z$-axis indicates the vertical direction in the driving space, $x$-axis indicates the horizontal direction. $f_{s0}$ is the intersection point of the LiDAR ray (red dashed line) and the ground. $f_{s1}, f_{s2}$, and $f_{s3}$ are the sampled free space points belonging to the ray path.}
    \label{fig:free_space}
\end{figure}

\textbf{Free space augmentation.}
In a point cloud, the free space -- traversed space by the ray that is not occupied by any obstacles -- is often neglected.
However, this information is also a result of the measurement and is critical for identifying the occupancy and visibility of the driving space. 
Therefore, we augment the point cloud data by sampling points from the LiDAR ray paths and call these points free space points $f_{si}$, where $i\in\mathbb{N}$.
As exemplified in Figure \ref{fig:free_space}, 
a LiDAR (orange cylinder) casts a ray (red line) that hits the surface of the ground at point $fs_0$ and only records this reflected point into the point cloud. 
Over the ray path, we sample free space points $f_{si}$ in a limited distance $d_{fs}$ from the hit point $f_{s0}$ with a large step $s_{fs}$. 
In order to constrain the computational overhead, we only sample points in the region of a limited height, \ie~$z\leq h_{fs}$, over the ground (blue area) where it is critical for driving. 
Finally, these points are down-sampled again by voxel down-sampling with a given voxel size of $v_{fs}$ to obtain evenly and sparsely distributed free space points.
These augmented free space points are then added to the original point cloud by setting their intensity value $\mathfrak{i}=-1$ as the indicator.
\label{sub:dataagumentation}

\textbf{Geometric augmentation.} We augment the point cloud data by randomly rotating, flipping along $x$- and $y$-axis and then scaling the geometric size of the point clouds. Also, we add a small Gaussian noise $[\delta x, \delta y, \delta z]$ to all the points. This augmentation helps increase the robustness of the model against domain shift during testing.

\subsection{Evaluation Metrics}
We use Intersection over Union (IoU) as a metric to evaluate the overall prediction performance and the calibration plot to analyze the quality of the predictive uncertainty. 
Although our proposed model GevBEV can generate BEV maps of any resolution, we only evaluate the predicted BEV maps of resolution of $0.4\,m$ so that it can be compared with other methods only generating the BEV maps of this predefined prediction resolution.

\paragraph{\textbf{IoU}} We report both the result overall perception areas in the predefined prediction range ($IoU_\text{all}$) and the result over the observable areas ($IoU_\text{obs}$), which is in line with the concept that the prediction is reliable only when it is conducted over the observed areas. 
Since our model is designed not to conduct predictions over non-observable areas, all the points in non-observable areas are regarded as false when calculating $IoU_\text{all}$. 
 Mathematically, a point $\mathbf{x}_j$ is observable if $\exists i \in \{i|\parallel \mathbf{x}_i - \mathbf{x}_j \parallel_2 <\nu\}$, meaning a point is observable if it is in the range $\nu=2\,m$ of any center points. The IoUs over these observable areas are calculated by Eq~\eqref{eq:iou_obs1} - \eqref{eq:iou_obs2}. 
\begin{align}
    \mathbf{X} &= \{\mathbf{x}_j |u_j < u_\text{thr}\}, \\ \label{eq:iou_obs1}
    \mathbf{X}^\text{fg} &= \{\mathbf{x}_j | \hat{p}_j^\text{fg} > \hat{p}_j^\text{bg}, \mathbf{x}_j\in\mathbf{X}\},\\
    \mathbf{Y} &= \{\mathbf{x}_j|\text{lbl}(\mathbf{x}_j)=\text{fg}\},\\
    \mathbf{Y}^\text{fg} &= \{\mathbf{x}_j|\mathbf{x}_j\in\mathbf{Y}, \mathbf{x}_j\in\mathbf{X}\},\\
    IoU &= \frac{|\mathbf{X}^\text{fg} \cap  \mathbf{Y}^\text{fg}|}{|\mathbf{X}^\text{fg} \cup  \mathbf{Y}^\text{fg}|}. \label{eq:iou_obs2}
\end{align}
We only evaluate the results of points $\mathbf{X}$ that have uncertainties under the uncertainty threshold $u_\text{thr}$. $\mathbf{X}^\text{fg}$ is the set of positive predictions -- points classified as foreground by its predictive classification probabilities.
$\text{lbl}(\mathbf{x}_j)$ is the class label of point $\mathbf{x}_j$, hence $\mathbf{Y}$ is the set of true positive points. $\mathbf{Y}^\text{fg}$ is the set of all positive points in the ground truth. Then the IoU is calculated between the true positive and ground truth positive points.

\paragraph{\textbf{Calibration plot}} We plot classification accuracy versus uncertainty to show their desired correlation. 
Our model generates uncertainty for the classification of each point, which enables an overall evaluation of all classes.  
However, the unbalanced number of samples for different classes will lead to a biased evaluation. 
For example, the classification of a point has a high uncertainty due to the lack of evidence from its neighbors, while it may end up with high classification accuracy because its neighbors belong to a dominant class.
To avoid this biased evaluation, each sample is weighted by the ratio of the total number of samples in that particular class.
Then, we divide the uncertainty $u\in[0, 1]$ into ten bins, and each bin has an interval of $0.1$.
Subsequently, we calculate the weighted average classification accuracy of all the samples in that uncertainty interval. 
A perfect calibration plot is shown by a diagonal line indicating the highest negative correlation between classification accuracy and uncertainty, \ie~high accuracy is associated with low uncertainty.

\subsection{Baseline and Comparative Models}
\label{subsec:sotamodels}
We evaluate the effectiveness of our proposed model GevBEV compared to both state-of-the-art co-perception camera- and lidar-based models on the co-perception benchmarks OPV2V and V2V4Real in both simulated and real driving scenarios, namely \textit{FCooper}~\cite{Fcooper}, \textit{AttFuse}~\cite{xu2022opencood}, \textit{V2VNet}~\cite{wang2020v2vnet}, \textit{DiscoNet}~\cite{li2021learning}, \textit{V2XViT}~\cite{xu2022v2x} and \ie~\textit{CoBEVT}~\cite{xu2022cobevt}.
Unlike our GevBEV, none of these models listed above provide uncertainty estimation for perception and distilling the essential information communicated to the ego CAV. 
Hence, we only compare GevBEV with those models on the perception performance. 
They are not further compared with GevBEV in terms of CPM size for the V2V communication in the co-perception application.

Moreover, we conducted a series of ablation studies to analyze the efficacy of the proposed modules of GevBEV.
\begin{itemize}[noitemsep]
    \item \textit{BEV} is the proposed model with the point-based spatial Gaussian and the evidential loss $\mathcal{L}_\textit{edl}$ removed, turning our model from a probabilistic model into a deterministic model. It uses cross-entropy to train the corresponding heads to classify points of the BEV maps. We treat this model as our baseline model.
    \item \textit{EviBEV} only has the point-based spatial Gaussian removed. It still uses $\mathcal{L}_\textit{edl}$ to train the distribution head.    
    \item \textit{GevBEV\textsuperscript{--}} is our proposed model but trained without free space augmentation.
\end{itemize}

\subsection{Implementation Details}
\label{subsubsec:implementation}
In all our experiments, we set the input voxel size to \SI{0.2}{m} to balance between computational overhead and performance. The free space points are sampled with the configuration $h_{fs}=-1.5\,m, d_\textit{fs}=1.5m, s_{fs}=7.5\,m$ on OPV2V. However, we increase $s_{fs}$ to $9m$ for V2V4Real dataset because it has a longer detection range in $x$-direction. During the geometric augmentation, the point cloud coordinates are scaled randomly in the range of $[0.95, 1.05]$ and then added with normally distributed $\mathcal{N}(0, 0.2)\,m$ noise. For the detection head, we generate two reference anchors at each observation point on the BEV map of stride four. These two anchors have the same size, $[l, w, h]=[4.41, 1.98, 1.64]\,m$ for OPV2V and  $[l, w, h]=[3.90, 1.60, 1.56]\,m$ for V2V4Real. However, they have different angles, $0^{\circ}$ and $90^{\circ}$, respectively. The anchors that have an {IoU} with the ground truth bounding boxes over 0.4 are regarded as positive, and those under 0.2 as negative; other anchors are neglected for calculating the loss during training. The negative samples are many more than the positive ones, leading to difficulties in the training process. Hence, we only randomly sample 512 negative samples for training.

We have trained the whole network from scratch for 50 epochs with weight decay of $0.01$ using the Adam optimizer~\cite{kingma2014adam} parameterized with $\beta=[0.95, 0.999]$ and $ \gamma=0.1$. We use multi-step learning-rate scheduler with a starting learning rate of $lr = 10^{-3}$ and two milestones at epoch 20 and 45. The learning rate decreases at each milestone by a factor of 0.1.

For the sampling generators, we use $n_\text{tgt}=10$ for the road head and $n_\text{tgt}=1$ for the objects head. The resolution of voxel down-sampling of the road head is set to \SI{0.4}{m}, and $N_\text{tgt}$ is set to 3000.
For the objects head, all the generated target points with a minimum distance of less than \SI{4}{m} to any ground truth boxes' edges are kept. From the background, we sample  $N_\text{tgt} = 50 \cdot n_\text{gt}$ points as negative samples, where $n_\text{gt}$ is the number of the ground truth bounding boxes. For these sampled target points, the evidences are drawn by Eq.~\eqref{eq:evi} in a maximum distribution range of $\nu=2\,m$. The parameters mentioned above are all set empirically, and the code and the trained model is released at \url{https://github.com/YuanYunshuang/GevBEV} for reproduction.

\subsection{Results}
\label{sec:results}
\subsubsection{Quantitative Analysis}
\noindent\textbf{Compared to state-of-the-art models.} Our proposed GevBEV model is benchmarked with the state-of-the-art models for co-perception on the simulated OPV2V dataset~\cite{xu2022opencood} and the real dataset V2V4Real \cite{xu2023v2v4real}.
Table \ref{tab:sota} lists the results for road and dynamic object segmentation.
It can be seen that the models conducted on camera data is inferior to LiDAR data.
This is because LiDAR data provides more accurate 3D information, which is essential for the projection to a BEV map for the perception task.
GevBEV outperforms all the other models, including the models that are conducted on the same LiDAR data as GevBEV for a fair comparison.
Compared to the runner up model CoBEVT on the OPV2V benchmark, our model with the distribution heads improves the IoU by 23.7\% for segmenting dynamic objects and 4.7\% for segmenting road surfaces. On the V2V4Real Benchmark, surprisingly, our model improves the IoU by 54.8\% compared to V2XViT.
These improvements indicate that the point-based spatial Gaussian effectively provides smoother information about each surface point's neighborhood, leading to more accurate results on both benchmarks. 
Besides, our proposed sampling method for training is more robust against the errors in ground truth. 
This leads to a remarkable improvement on the real dataset V2V4Real that contains inaccurate labels in real-world driving.

\begin{table}[t!]
\centering
\setlength{\tabcolsep}{2pt}
\begin{tabular}{l|cc|cc|c}
\toprule
\multirow{2}{*}{Model} & \multicolumn{2}{c|}{Modality} & \multicolumn{2}{c|}{OPV2V} & \multicolumn{1}{c}{V2V4Real} \\ 
         &Camera & LiDAR& \quad Road  \quad   & \quad Object \quad   & \quad Object  \quad   \\ \midrule
  AttFuse \cite{xu2022opencood} & \checkmark &  & 60.5 & 51.9 & - \\
  V2VNet\cite{wang2020v2vnet} & \checkmark &  & 60.2 & 53.5 & - \\
  DiscoNet \cite{li2021learning} & \checkmark &  & 60.7 & 52.9 & - \\
  CoBEVT \cite{xu2022cobevt} & \checkmark &  & 63.0 & 60.4 & - \\
  Fcooper \cite{Fcooper} & & \checkmark & 70.3 & 52.1 & 25.9 \\
  AttFuse \cite{xu2022opencood}&  & \checkmark & 75.3 & 52.0 & 25.5 \\
  V2XViT \cite{xu2022v2x} & & \checkmark & 75.0 & 50.4 & \underline{29.9} \\
  CoBEVT \cite{xu2022cobevt} & & \checkmark & \underline{75.9} &\underline{52.3} & 29.6 \\
  GevBEV (ours) & & \checkmark & \textbf{79.5} & \textbf{74.7} & \textbf{46.3} \\
\bottomrule
\end{tabular}
\caption{Comparison with the state-of-the-art models on OPV2V and V2V4real dataset. Best values are highlighted in boldface and the second best values are underlined.}
\label{tab:sota}
\end{table}
However, the real-world driving scenarios post more challenges for the co-perception task by inevitably introducing localization errors, as indicated by the large performance gap on object segmentation between OPV2V (74.7\%) and V2V4Real (46.3\%), 
Also, the V2V4Real dataset only provides the communication between two connected vehicles, fewer than that of the simulated OPV2V dataset. 
Hence, the perception performance on V2V4Real is much worse than that tested on OPV2V.



\vspace{12pt}
\begin{figure*}[t!]
  \centering
    \includegraphics[width=0.4\linewidth]{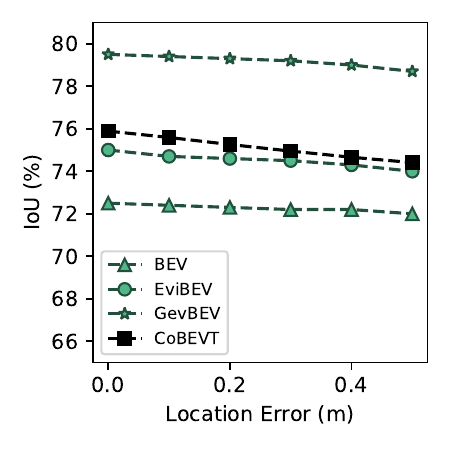}
    \includegraphics[width=0.4\linewidth]{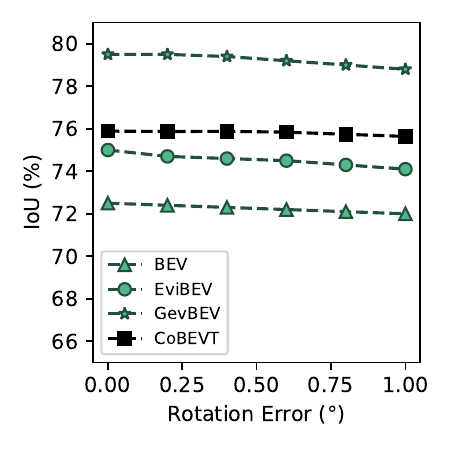}
    \caption{OPV2V road segmentation result with localization noise.}
    \label{fig:loc_err_opv2v_road}
\end{figure*}

\begin{figure*}[t!]
  \centering
    \includegraphics[width=0.4\linewidth]{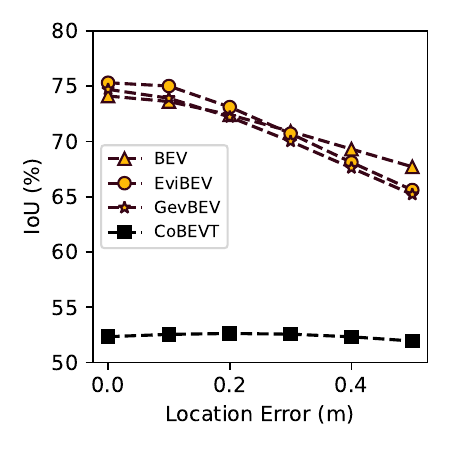}
    \includegraphics[width=0.4\linewidth]{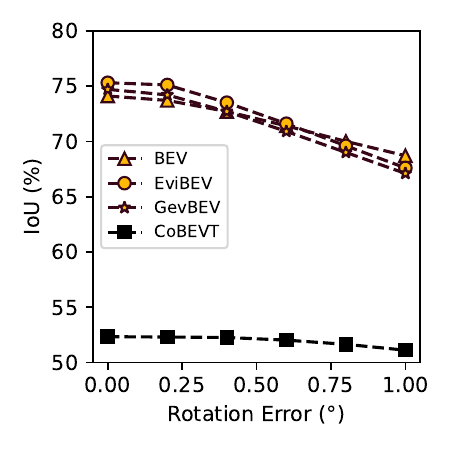}
    \caption{OPV2V object segmentation result with localization noise.}
    \label{fig:loc_err_opv2v_object}
\end{figure*}

\begin{figure*}[t!]
  \centering
    \includegraphics[width=0.4\linewidth]{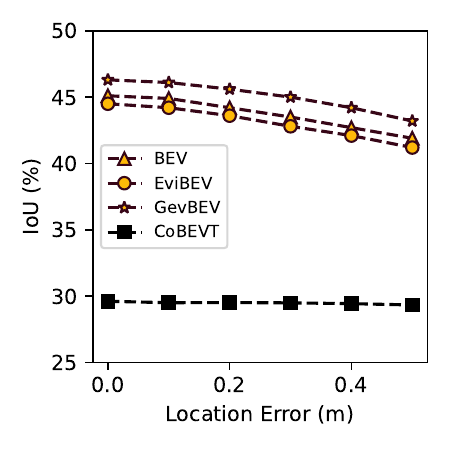}
    \includegraphics[width=0.4\linewidth]{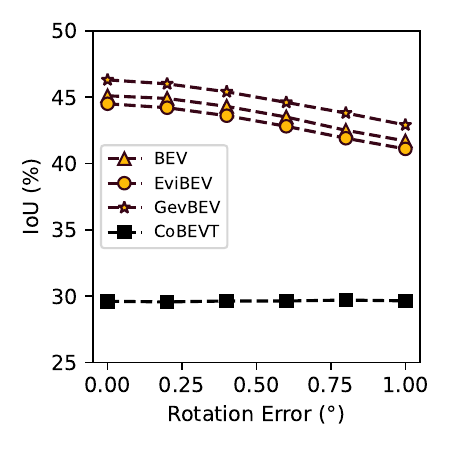}
    \caption{V2V4Real object segmentation result with localization noise.}
    \label{fig:loc_err_v2vreal_object}
\end{figure*}
\noindent\textbf{Sensitivity to localization noise.}
In the previous experiments of the simulated data, we assumed perfect localization information.
In order to evaluate the simulation of the results to localization noise, we introduce localization errors generated by normal distributions with a standard deviation ranging from 0 to 0.5 meters for position and from 0 to 1 degrees for orientation. Figure \ref{fig:loc_err_opv2v_road} demonstrates that all models experience slight performance declines in road surface segmentation as the location and rotation errors increase. Still, our proposed GevBEV model outperforms CoBEVT, maintaining the best performance with a margin of approximately 4\% across all error configurations.
In contrast to the segmentation performance, objects exhibit higher sensitivity to localization errors due to their smaller size. 
In this setting, our model still outperforms the runner up model CoBEVT on both the OPV2V and V2V4Real datasets, as depicted in Figures \ref{fig:loc_err_opv2v_object} and \ref{fig:loc_err_v2vreal_object}.
This indicates that our proposed approach is more robust than CoBEVT to cope with localization noise.

\vspace{12pt}
\noindent\textbf{Ablation study.} Table~\ref{tab:ablation} shows the performances of the ablative models.
In general, the baseline model BEV without the point-based spatial Gaussian (G.s.) and the evidential loss $\mathcal{L}_\textit{edl}$ is inferior to the other models. This indicates that this conventional deterministic model trained by optimizing the cross-entropy is not as good as the probabilistic models. In contrast, by modeling each point with a spatial continuous Gaussian distribution, we are able to close the gaps caused by the sparsity of point clouds and generate smoother BEV maps.
EviBEV with the point-based spatial Gaussian performs better than the baseline for the surface measured by the {IoUs} for \textit{all} and the \textit{observed} areas.
However, its performances for objects are slightly degraded.

\begin{table}[ht]
 \centering
 \setlength{\tabcolsep}{0.5pt}
\begin{tabular}{l|ccc|cc|c|cc|c}
\toprule
\multirow{2}{*}{Model} & \multicolumn{3}{c|}{Modules} & \multicolumn{2}{c|}{OPV2V(all)} & \multicolumn{1}{c|}{V2V4Real(all)}  & \multicolumn{2}{c|}{OPV2V(obs)} & \multicolumn{1}{c}{V2V4Real(obs)} \\ 
         & G.s.  &$\mathcal{L}_\textit{edl}$     & Ag.    & Rd        & Obj      & Obj         & Rd        & Obj      & Obj         \\ \midrule
BEV      &        &          &\checkmark& 72.5        &  74.1   & 45.1   & 76.1           & 75.8    &46.1     \\
EviBEV   &        &\checkmark&\checkmark& 75.0         & \textbf{75.3}   & 44.5     & 78.3         & \textbf{76.3} & 45.3       \\
GevBEV$^-$&\checkmark&\checkmark&         & 59.7         & 73.1    & 46.0     & 62.5          & 73.2    & 46.7   \\
GevBEV  &\checkmark&\checkmark&\checkmark& \textbf{79.5 }         & 74.7   & \textbf{46.3}     & \textbf{83.1}          & 76.1   & \textbf{46.9}     \\ \bottomrule
\end{tabular}
\caption{Ablation study of the proposed modules. All IoU results are measured in percentage. Best values are highlighted in boldface. G.s.: point-based spatial Gaussian,  $\mathcal{L}_\textit{edl}$: evidential loss, Ag.: free space augmentation; Rd: road, Obj: object.}
    \label{tab:ablation}
\end{table}

Moreover, BEV and EviBEV perform worse than CoBEVT \cite{xu2022cobevt} on $IoU_\text{all}$ of the road head, as shown in Table~\ref{tab:sota}. 
This is because our frameworks are based on fully sparse convolution networks, which do not operate on unobserved areas. In order to facilitate comparison with dense convolution models, certain road surfaces in our frameworks are considered inadequately observed and thus treated as false predictions when calculating $IoU_\textit{all}$. Unlike GevBEV, both our BEV and EviBEV models have additional areas designated as unobserved due to the absence of Gaussian tails. Consequently, this leads to lower IoU values.
However, it is worth mentioning that the object classification of the ablative models significantly outperforms the runner up model, CoBEVT. This can be attributed to two factors. Firstly, we carefully control the network in our model to expand coordinates in a specific range and only make predictions over the observable areas.
The coordinate expansion module can cover most of the object areas so that these areas will be given a prediction rather than being treated as unobserved. 
Secondly, thanks to the benefits of dynamic sampling from continuous driving space during training, our model shows a tendency to be cautious when making positive predictions for vehicle points. This cautious approach allows us to capture edge details of the vehicles more effectively, enhancing the overall object classification performance.

From the comparison between GevBEV$^-$ and GevBEV, the improved {IoUs}, especially for roads, indicate that the free space augmentation (Ag.) provides an explicit cue to the unoccupied space along the ray paths and improves the detection performance. 
Also, this module plays an important role in mitigating the problem of sparsity for point clouds and largely improves the prediction performance.
Overall, GevBEV outperforms the ablative models in all the measurements, which validates the efficacy of each proposed module.

\subsubsection{Visual Analysis}

\begin{figure*}[t]
  \centering
    \includegraphics[width=1.0\linewidth]{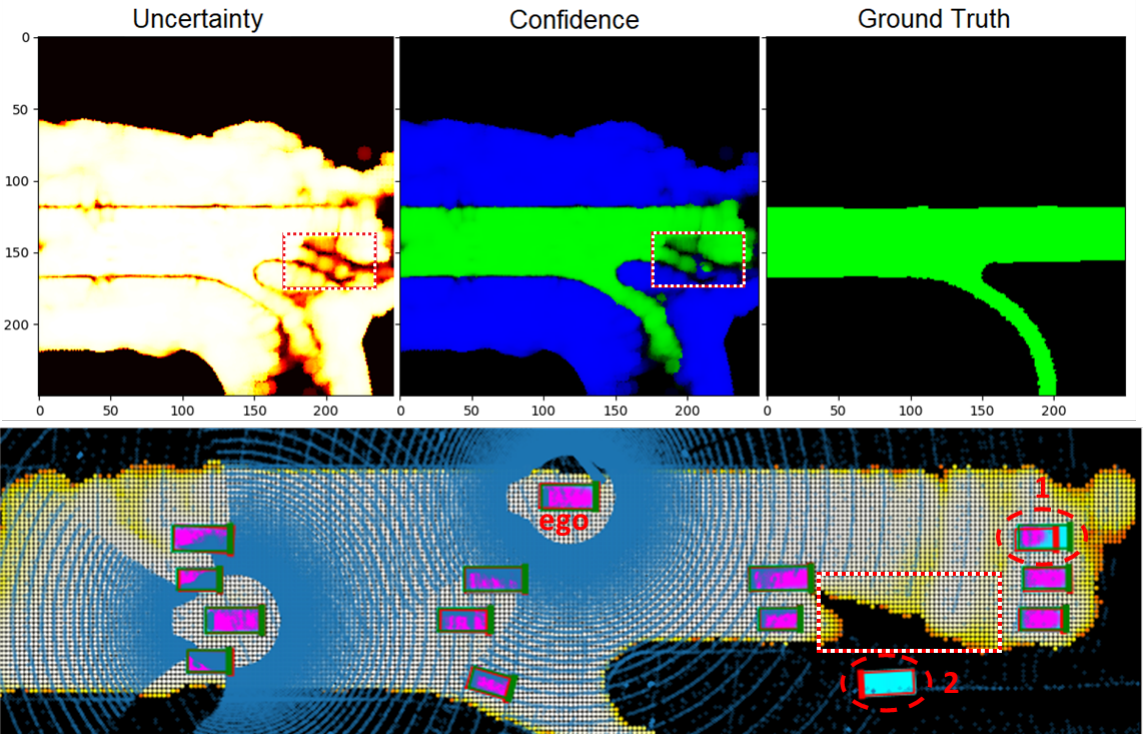}\\

    \caption{Results of the GevBEV maps. In the \textit{Uncertainty map}, lower uncertainty is presented in a lighter color. In the \textit{Confidence map}, road confidence is indicated in green and background confidence in blue with different intensities. The bottom sub-figure shows the more detailed detection results in a larger extent. Specifically, the original input point cloud is denoted by blue points, road points are highlighted by a light color if their uncertainties are under the threshold of 0.7, and the objects are shown in bounding boxes with red color indicating the detection, and with green color the corresponding ground truth, respectively. The thick bar in the front of the bounding boxes denotes the driving direction. Moreover, those bounding boxes are filled with the point confidences drawn from the objects head, where magenta points are associated with high confidence, while cyan indicates the opposite.}
    \label{fig:example_result}
\end{figure*}

\noindent\textbf{Holistic BEV maps for autonomous driving.} 
With our proposed probabilistic model, we generate the GevBEV maps and visualize the results in a complex driving scenario in Figure \ref{fig:example_result}. From left to right, the three sub-figures in the first row show the results of uncertainty, classification confidence, and the ground truth of the road surface. In the uncertainty map, a lighter color indicates lower uncertainty, whereas black areas are regarded as non-observable. Correspondingly, the confidence map gives the confidence score for both foreground (road surface) and background. The bottom sub-figure is the detailed detection results of both road surface and objects overlaid in one figure. Only the points that are classified as roads with uncertainty under the threshold of 0.7 are highlighted in the light color in the bottom layer to show the situation of the drivable area. The predicted and corresponding ground truth bounding boxes are plotted in red and green colors, respectively. Moreover, the bounding boxes are filled with the point confidences drawn from the objects head, where magenta points are associated with high confidence, while cyan the opposite.

As shown in Figure \ref{fig:example_result}, the GevBEV maps are regarded as holistic BEV maps providing a reliable information source to support AVs for decision-making.
First, this information can be sourced back to the original measurement points. For example, due to occlusions and long detection range, the area marked with a dashed line red rectangle is an area where the ego CAV is not certain about.
If the ego CAV needs to drive into this uncertain area, it should either request the missing information from other CAVs or slow down waiting for the clearance of the uncertain area. 
Second, the object's head generates evidence of the areas that might be occupied by vehicles. 
Therefore, they are a reliable and explainable clue to validate the bounding box detection. 
For example, as shown in the bottom sub-figure, most predicted boxes are well aligned with the ground truth except those two marked with red ellipses. The vehicle in ellipse 1 is only partially observed; there is not enough evidence to support this detection. Similarly, there is nearly no evidence for false positive detection in ellipse 2. Therefore, this detection can be simply removed. 

\begin{figure}[t!]
  \centering
    \includegraphics[width=0.55\linewidth]{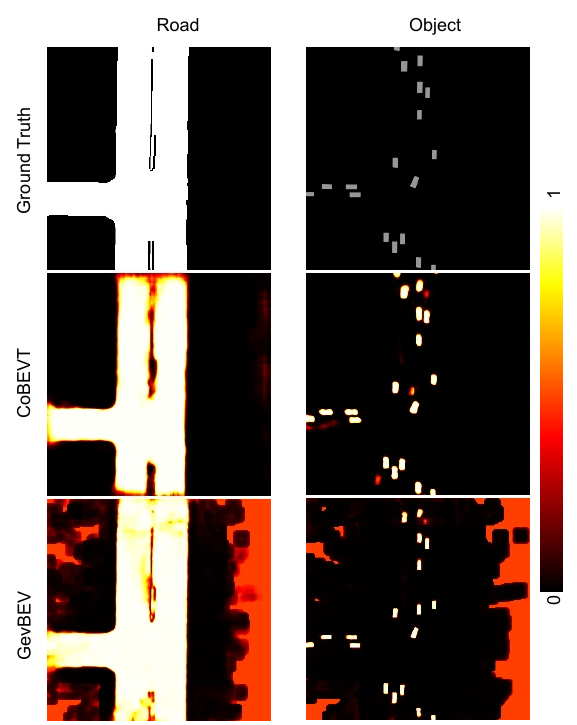}
    \caption{The comparison of classification confidences between GevBEV and CoBEVT on the OPV2V dataset.}
    \label{fig:bev_opv2v}
\end{figure}

\begin{figure}[t!]
  \centering
    \includegraphics[width=0.99\linewidth]{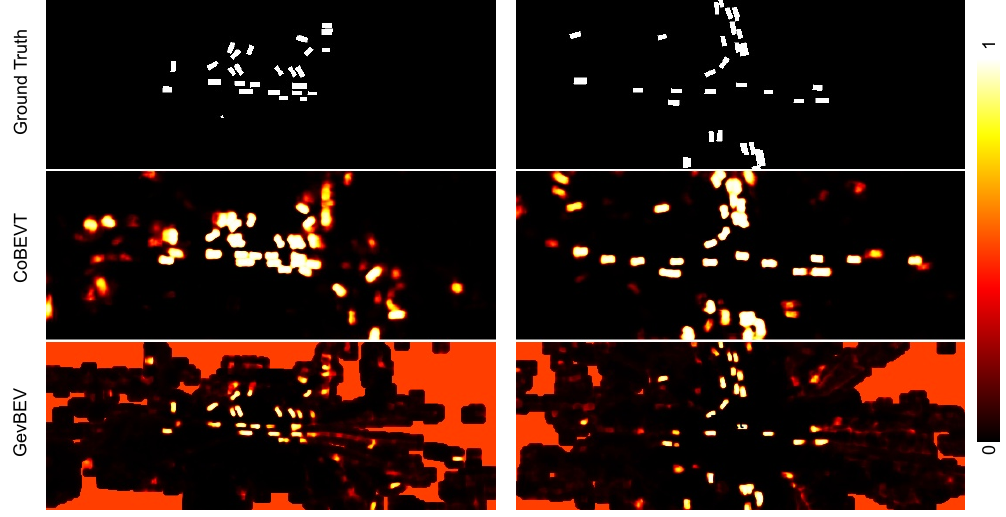}
    \caption{The comparison of classification confidences between GevBEV and CoBEVT on the V2VReal dataset.}
    \label{fig:bev_v2vreal}
\end{figure}

\noindent\textbf{Comparison to baseline.} Figures \ref{fig:bev_opv2v} and \ref{fig:bev_v2vreal} illustrate the classification confidences, with the color scale transitioning from dark to light, representing confidence values from 0 to 1. In the right column of Figure \ref{fig:bev_opv2v} and the bottom row of figure \ref{fig:bev_v2vreal}, the red color indicates that our model retains information about unobserved areas. In the absence of observations, the model remains unbiased towards predicting either the foreground or the background class. Additionally, our model exhibits a tendency to produce more refined details for both road and vehicle edges. Notably in Figure \ref{fig:bev_v2vreal}, CoBEVT tends to generate more false positive predictions, and in some cases, vehicles even appear merged together. In contrast, our model accurately separates all vehicles, thanks to its precise edge description.

\begin{figure}[t!]
  \centering
    \includegraphics[width=0.7\linewidth]{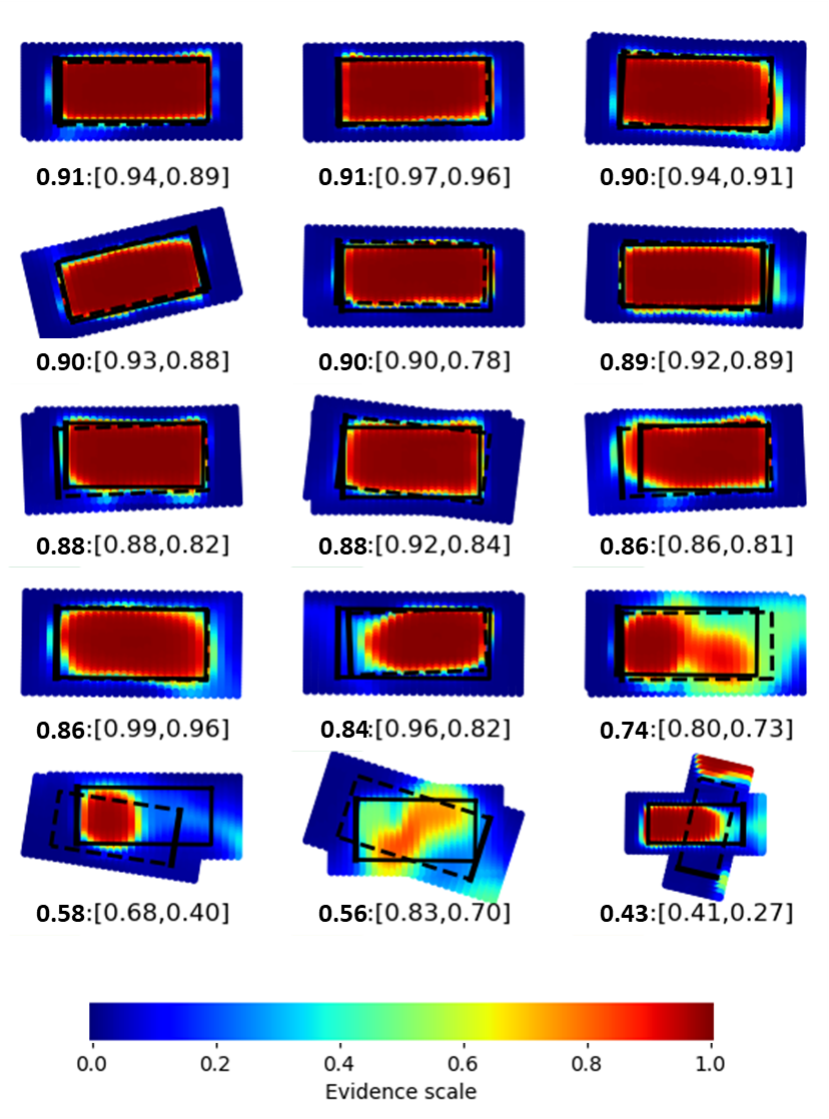}
    \caption{Object point distribution. The dished line bounding boxes denote the detection, and the solid line bounding boxes denote the corresponding ground truth. The thick bar of each bounding box denotes the driving direction. The statistics under each sub-figure denote the average evidence and [JIoU, IoU].}
    \label{fig:box_distr}
\end{figure}

\vspace{12pt}
\noindent\textbf{Evidence of object point distribution.}
Furthermore, we use the average evidence score to better show the relations between the quality of detection and the corresponding object point distributions.
Inspired by the work from Wang \etal~\cite{Wang2020InferringSU}, in addition to the normal IoU over the detection areas, we also leverage the generated uncertainty to calculate the JIoU. JIoU is a probability version of IoU that better evaluates the probabilistic features of object detection. Slightly different to~\cite{Wang2020InferringSU} that defines JIoU as the IoU between the probability mass covered by the detection and the ground truth bounding boxes, we define it as the IoU between the sum of the evidences in the detected bounding boxes and the sum of all the evidence masses describing this object.
Then, we calibrate the average evidence score for a single detected bounding box, which is the mean of the drawn confidences of points inside the bounding boxes. This JIoU ensures low evidence score when the predicted box is fully filled with strong evidences but does not cover all evidences that describe this object. 

The results in Figure \ref{fig:box_distr} demonstrate that the average evidence score is very well related to the quality of the detection.
As shown in the last row, worse detection tends to have less evidence inside the predicted bounding boxes.
Moreover, JIoU reveals the alignment of the prediction and ground truth bounding boxes over the evidence masses. 
Objects without enough clues from the measurements are hard to define a perfect deterministic ground truth, even by manual labeling, so does fairly judge the model's prediction based on this. In such cases, both detection and ground truth bounding boxes have low evidence coverage over the GevBEV map. This leads to a smaller probabilistic union between these two boxes, hence a higher JIoU is derived. Compared to IoU, this is more reasonable as JIoU decouples the imperfectness of the model and the measurement. For example, the detected bounding box in the lower-left subfigure has low IoU but relatively high JIoU because there are too few evidences caused by the lankness of the observation points -- measurement imperfectness. In contrast, compared to the ground truth, the detection in the lower-right subfigure has enough evidences but has both low JIoU and IoU, indicating that the inferior detection is not due to the measurement but the model's limited performance. 
 
\begin{figure}[t]
  \centering
    \includegraphics[width=\linewidth]{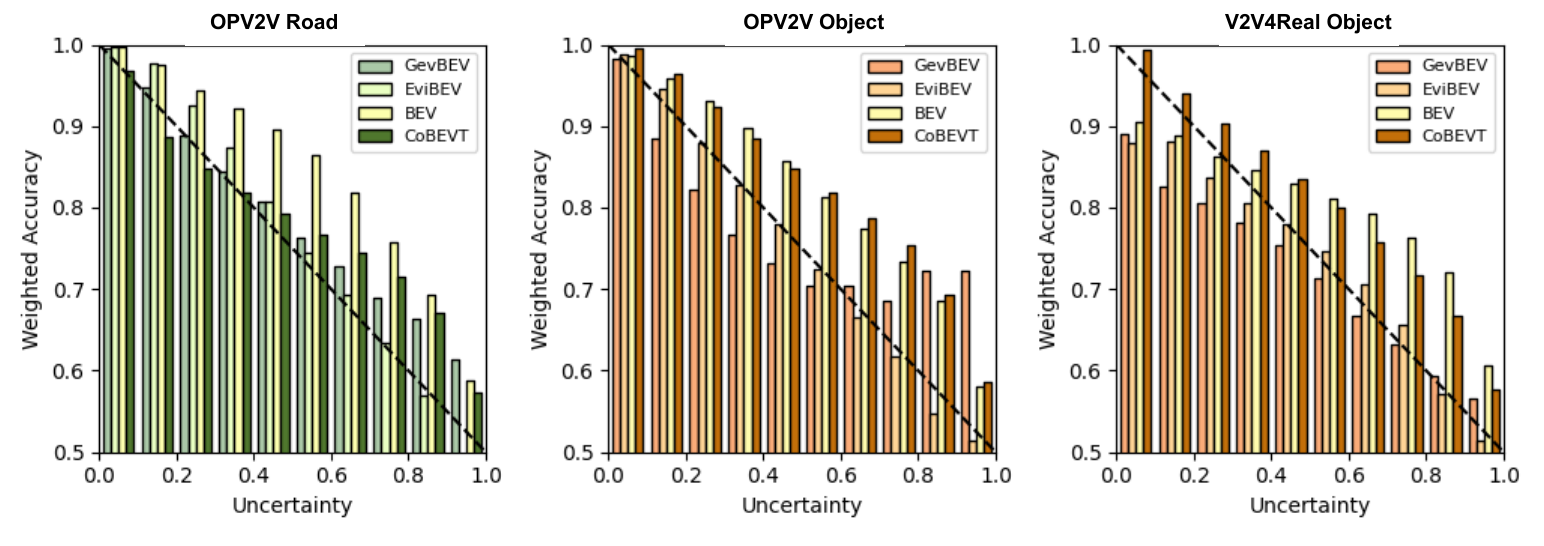}
    \caption{Calibration plots by different models. The perfect calibration line is indicated by the diagonal dashed line.}
    \label{fig:uncq}
\end{figure}
\begin{table}[th]
\centering

\setlength{\tabcolsep}{2pt}
\begin{tabular}{l|cc|c|c}
\toprule
\multirow{2}{*}{Model} & \multicolumn{2}{c|}{OPV2V} & \multicolumn{1}{c|}{V2V4real} & \multirow{2}{*}{ Average }\\ 
         & Road  \quad   & \quad Object \quad   & \quad Object  \quad &   \\ \midrule
  BEV   & 0.095 & 0.072 & 0.076 & 0.081\\
  EviBEV   & \textbf{0.031}  & \textbf{0.010} & \textbf{0.030} & \textbf{0.023} \\
  CoBEVT   & \underline{0.044} & 0.075 & 0.056 & 0.058\\
  GevBEV  & \underline{0.044} & \underline{0.066} & \underline{0.040} & \underline{0.050}\\
\bottomrule
\end{tabular}
\caption{Average offset of calibration plot to the perfect calibration line. Best values are highlighted in boldface and the second best values are underlined.}
\label{tab:uncq}
\end{table}
\vspace{12pt}
\noindent\textbf{Desired confidence level with calibration plot.}
We use the calibration plot (Figure \ref{fig:uncq}) and the average offsets (Table \ref{tab:uncq}) between this plot and the perfect uncertainty-accuracy line (dashed black line) as a summary to analyze the quality of uncertainty generated by the baseline model CoBEVT and  BEV, the EviBEV model with $\mathcal{L}_\textit{edl}$ loss, and our complete probabilistic model GevBEV. 
Since the baseline model only generates Softmax scores for each class, we convert the scores into entropy to quantify the uncertainty and compare it to the other two models with the evidential uncertainty based on a Dirichlet distribution.
As revealed in Figure \ref{fig:uncq}, GevBEV and EviBEV demonstrate better confidence plots to the perfect calibration line (indicated by the diagonal dashed line) than the baseline model BEV and CoBEVT for both road and object classfication. 
The two baseline models seem to overestimate the uncertainty than the other two models, which affirms our concerns that the deterministic model, without particularly accounting for uncertainties, may end up generating less trustworthy scores for making driving decisions.
The results, on the other hand, show that assuming a Dirichlet distribution of the point class of the BEV map can provide more reliable probabilistic features for the map and, therefore, is safer to use in AV perception systems.

Interestingly, the uncertainty quality of GevBEV is worse than that of EviBEV, especially for object classification. This might be caused by the saturation of the summation of the evidences contributed by the neighboring center points. 
Moreover, the highly uncertain points from the objects head of GevBEV tend to be underconfident. We conjecture that some vehicles are only observed partially because of occlusion.
Our limited coordinate expansion ($1.2\,m$) is only able to cover parts of these vehicles. Therefore, only the distribution tiles of these expanded center points can cover the rest of the vehicle body.
This then may lead to a high uncertainty but a high positive rate.
Despite the uncertainty of GevBEV being slightly more conservative than that of EviBEV, still, as shown by the higher {IoUs} in Table~\ref{tab:ablation}, the learned spatial Gaussian distribution generates smoother BEV maps and draws classification distribution of any points in the continuous BEV 2D space.

\subsection{The Application of GevBEV for Co-Perception}
\begin{figure}[t!]
  \centering
    \includegraphics[width=\linewidth]{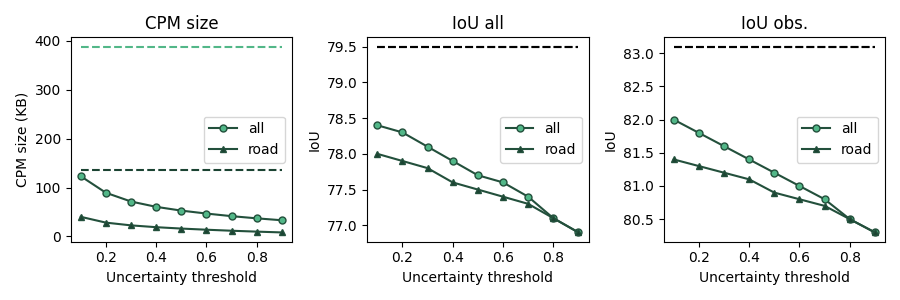}\\
    \includegraphics[width=\linewidth]{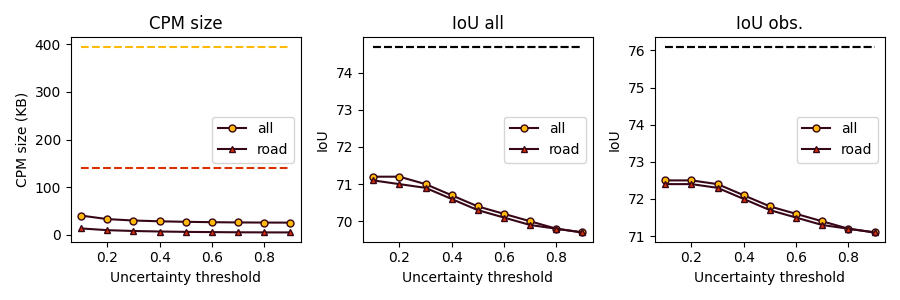}\\
    \caption{Comparison of different CPM sharing strategies for co-perception. The first row shows the results of the road head (greenish) and the second the objects head (orangish).}
    \vspace{-12pt}
    \label{fig:cpm_size}
\end{figure}

The generated evidential GevBEV maps with different uncertainty thresholds are used to select the information communicated among CAVs. 
The CPM sizes before and after the uncertainty-based information selection with different uncertainty thresholds, as well as the corresponding IoUs of the classification results over all areas in the perception range ($IoU_\text{all}$) and over the observable areas ($IoU_\text{obs}$), are plotted in Figure \ref{fig:cpm_size}.  
The first row shows the results of the road head, and the second the objects head. As discussed in section \ref{sec:co_perception}, we conducted experiments with two CPM sharing strategies, one for sharing the masked evidence maps of all perception areas (light green and orange) and one only for the road areas constrained by an existing HD map (dark green and orange). The dashed lines are the baselines of sharing CPMs without information selection, which are shown as horizontal lines over different uncertainty thresholds. 

The plots in the first column show that the CPM sizes have been reduced evidently after the information selection at all uncertainty thresholds compared to the corresponding baselines. For example, when all perception areas (light green) are considered for sharing, the CPM size for the road head dropped by ca. 87\% from 388\,KB to 52\,KB at the uncertainty threshold of 0.5. Correspondingly, the $IoU_\text{all}$ and $IoU_\text{obs}$ only dropped by ca. 2\%. In the same configuration, the CPM sizes for the object detection dropped ca. 93\%, from 395KB to 27KB, while the IoUs dropped ca. 4\%. By only considering the road areas for sharing, CPM sizes can be further reduced to about 16\,KB for the road head and 6\,KB for the objects head at the uncertainty threshold of 0.5, as the dark green and orange solid lines shown with only an IoU drop within 0.5\%. According to the V2V communication protocol \cite{v2v_protocol}, without considering other communication overhead, the data throughput rate can achieve 27\,Mbps. Therefore, the time delay for sending the CPMs of both heads has dropped from ca. 28\,ms to 0.8\,ms by the data selection based on our GevBEV maps from road areas.
These significantly reduced time delays are critical for real-time V2V communication.

\section{Conclusion}
In this paper, we proposed a novel method to interpret driving environments with observable Gaussian evidential BEV maps. These maps interpret the LiDAR sensory data in a back-traceable manner that each prediction is supported by evidences provided by the original observation points.
Moreover, we designed a probabilistic classification model based on U-Net to generate statistics for this interpretation. This model assumes a spatial Gaussian distribution for each voxel of a predefined resolution so that any point in the continuous driving space can draw itself a Dirichlet distribution of the classification based on the evidences drawn from the spatial Gaussian distributions of the neighboring voxels.

We test our proposed GevBEV on benchmarks OPV2V and V2V4Real of BEV map interpretation for cooperative perception in simulated and real-world driving scenarios, respectively.
The experiments show that GevBEV outperforms the baseline of the image-based BEV map by a large margin. By analyzing the predictive uncertainty, we also proved that evidential classification can score the classification result in a less overconfident and better-calibrated manner than the deterministic counterpart of the same model. 
Furthermore, the spatial Gaussian distribution assigned to each observable point is also proven beneficial in closing the gaps of sparse point clouds with a controllable range and smoothing the BEV maps. 
By virtue of this spatial distribution, one can draw the Dirichlet classification result for any points in the continuous driving space. This probabilistic result can be used to make safer decisions for autonomous driving by its ability to quantify the uncertainty using the measurement evidences.

Although our model is only applied to the point cloud data of the co-perception scenario in this paper, it is straightforward to be used on other modalities of data sources or scenarios, such as images for co-perception, or satellite images and airborne point clouds in the field of remote sensing. We leave these applications for future work. 

\section{Acknowledgments}
This work was supported Deutsche Forschungsgemeinschaft (DFG, German Research Foundation) -- 227198829/GRK1931 and MSCA European Postdoctoral Fellowships under the 101062870 -- VeVuSafety project.

\bibliography{egbib.bib}

\begin{thebibliography}{10}
\expandafter\ifx\csname url\endcsname\relax
  \def\url#1{\texttt{#1}}\fi
\expandafter\ifx\csname urlprefix\endcsname\relax\def\urlprefix{URL }\fi
\expandafter\ifx\csname href\endcsname\relax
  \def\href#1#2{#2} \def\path#1{#1}\fi

\bibitem{lin2021local}
Y.~Lin, G.~Vosselman, Y.~Cao, M.~Y. Yang, Local and global encoder network for
  semantic segmentation of airborne laser scanning point clouds, ISPRS Journal
  of Photogrammetry and Remote Sensing 176 (2021) 151--168.

\bibitem{FengHWD22}
D.~Feng, A.~Harakeh, S.~L. Waslander, K.~Dietmayer, A review and comparative
  study on probabilistic object detection in autonomous driving, {IEEE} Trans.
  Intell. Transp. Syst. 23~(8) (2022) 9961--9980.

\bibitem{zhang2023perception}
Y.~Zhang, A.~Carballo, H.~Yang, K.~Takeda, Perception and sensing for
  autonomous vehicles under adverse weather conditions: A survey, ISPRS Journal
  of Photogrammetry and Remote Sensing 196 (2023) 146--177.

\bibitem{fang2022joint}
L.~Fang, Z.~You, G.~Shen, Y.~Chen, J.~Li, A joint deep learning network of
  point clouds and multiple views for roadside object classification from
  {LiDAR} point clouds, ISPRS Journal of Photogrammetry and Remote Sensing 193
  (2022) 115--136.

\bibitem{zang2017lane}
A.~Zang, R.~Xu, Z.~Li, D.~Doria, Lane boundary extraction from satellite
  imagery, in: Proceedings of the 1st ACM SIGSPATIAL Workshop on High-Precision
  Maps and Intelligent Applications for Autonomous Vehicles, 2017, pp. 1--8.

\bibitem{kitti}
A.~Geiger, P.~Lenz, R.~Urtasun, Are we ready for autonomous driving? the kitti
  vision benchmark suite, in: Conference on Computer Vision and Pattern
  Recognition (CVPR), 2012.

\bibitem{nuscenes}
H.~Caesar, V.~Bankiti, A.~H. Lang, S.~Vora, V.~E. Liong, Q.~Xu, A.~Krishnan,
  Y.~Pan, G.~Baldan, O.~Beijbom, nuscenes: A multimodal dataset for autonomous
  driving, in: Conference on Computer Vision and Pattern Recognition (CVPR),
  2020.

\bibitem{waymo}
P.~Sun, H.~Kretzschmar, X.~Dotiwalla, A.~Chouard, V.~Patnaik, P.~Tsui, J.~Guo,
  Y.~Zhou, Y.~Chai, B.~Caine, V.~Vasudevan, W.~Han, J.~Ngiam, H.~Zhao,
  A.~Timofeev, S.~Ettinger, M.~Krivokon, A.~Gao, A.~Joshi, Y.~Zhang, J.~Shlens,
  Z.~Chen, D.~Anguelov, Scalability in perception for autonomous driving: Waymo
  open dataset, in: Conference on Computer Vision and Pattern Recognition
  (CVPR), 2020, pp. 2443--2451.

\bibitem{Survey_unc}
J.~Gawlikowski, C.~R.~N. Tassi, M.~Ali, J.~Lee, M.~Humt, J.~Feng, A.~M. Kruspe,
  R.~Triebel, P.~Jung, R.~Roscher, M.~Shahzad, W.~Yang, R.~Bamler, X.~Zhu, A
  survey of uncertainty in deep neural networks, ArXiv abs/2107.03342 (2021).

\bibitem{fpvrcnn}
Y.~Yuan, H.~Cheng, M.~Sester, Keypoints-based deep feature fusion for
  cooperative vehicle detection of autonomous driving, IEEE Robotics and
  Automation Letters PP (2022) 1--1.

\bibitem{xu2022bridging}
R.~Xu, J.~Li, X.~Dong, H.~Yu, J.~Ma, Bridging the domain gap for multi-agent
  perception, arXiv preprint arXiv:2210.08451 (2022).

\bibitem{Cui2022CoopernautED}
J.~Cui, H.~Qiu, D.~Chen, P.~Stone, Y.~Zhu, Coopernaut: End-to-end driving with
  cooperative perception for networked vehicles, in: Conference on Computer
  Vision and Pattern Recognition (CVPR), 2022, pp. 17231--17241.

\bibitem{xu2022opencood}
R.~Xu, H.~Xiang, X.~Xia, X.~Han, J.~Li, J.~Ma, Opv2v: An open benchmark dataset
  and fusion pipeline for perception with vehicle-to-vehicle communication, in:
  IEEE International Conference on Robotics and Automation (ICRA), 2022.

\bibitem{Sensoy2018EvidentialDL}
M.~Sensoy, L.~Kaplan, M.~Kandemir, Evidential deep learning to quantify
  classification uncertainty, Advances in neural information processing systems
  (NeurIPS) 31 (2018).

\bibitem{xu2023v2v4real}
R.~Xu, X.~Xia, J.~Li, H.~Li, S.~Zhang, Z.~Tu, Z.~Meng, H.~Xiang, X.~Dong,
  R.~Song, H.~Yu, B.~Zhou, J.~Ma, V2v4real: A real-world large-scale dataset
  for vehicle-to-vehicle cooperative perception, in: The IEEE/CVF Computer
  Vision and Pattern Recognition Conference (CVPR), 2023.

\bibitem{xu2022cobevt}
R.~Xu, Z.~Tu, H.~Xiang, W.~Shao, B.~Zhou, J.~Ma, Cobevt: Cooperative bird's eye
  view semantic segmentation with sparse transformers, in: Conference on Robot
  Learning (CoRL), 2022.

\bibitem{overview_Jiao2019OD}
L.~Jiao, F.~Zhang, F.~Liu, S.~Yang, L.~Li, Z.~Feng, R.~Qu, A survey of deep
  learning-based object detection, IEEE Access 7 (2019) 128837--128868.

\bibitem{li2022domain}
J.~Li, R.~Xu, J.~Ma, Q.~Zou, J.~Ma, H.~Yu, Domain adaptive object detection for
  autonomous driving under foggy weather, in: IEEE Winter Conference on
  Applications of Computer Vision (WACV), 2022, pp. 612--622.

\bibitem{Kirillov2018PanopticS}
A.~Kirillov, K.~He, R.~B. Girshick, C.~Rother, P.~Doll{\'a}r, Panoptic
  segmentation, Conference on Computer Vision and Pattern Recognition (CVPR)
  (2018) 9396--9405.

\bibitem{qiu2022gfnet}
H.~Qiu, B.~Yu, D.~Tao, {GFN}et: Geometric flow network for 3d point cloud
  semantic segmentation, Transactions on Machine Learning Research (2022).

\bibitem{cvt_zhou2022}
B.~Zhou, P.~Kr{\"a}henb{\"u}hl, Cross-view transformers for real-time map-view
  semantic segmentation, in: Conference on Computer Vision and Pattern
  Recognition (CVPR), 2022.

\bibitem{bev_Loukkal2021Flatmobiles}
A.~Loukkal, Y.~Grandvalet, T.~Drummond, Y.~Li, Driving among flatmobiles:
  Bird-eye-view occupancy grids from a monocular camera for holistic trajectory
  planning, IEEE Winter Conference on Applications of Computer Vision (WACV)
  (2021) 51--60.

\bibitem{bev_Lu2019MonocularSO}
C.~Lu, M.~J.~G. van~de Molengraft, G.~Dubbelman, Monocular semantic occupancy
  grid mapping with convolutional variational encoder–decoder networks, IEEE
  Robotics and Automation Letters 4 (2019) 445--452.

\bibitem{bev_Pan2020CrossViewSS}
B.~Pan, J.~Sun, H.~Y.~T. Leung, A.~Andonian, B.~Zhou, Cross-view semantic
  segmentation for sensing surroundings, IEEE Robotics and Automation Letters 5
  (2020) 4867--4873.

\bibitem{li2022bevformer}
Z.~Li, W.~Wang, H.~Li, E.~Xie, C.~Sima, T.~Lu, Q.~Yu, J.~Dai, Bevformer:
  Learning bird's-eye-view representation from multi-camera images via
  spatiotemporal transformers, in: European Conference on Computer Vision,
  2022.

\bibitem{Yang2018PIXORR3}
B.~Yang, W.~Luo, R.~Urtasun, Pixor: Real-time 3d object detection from point
  clouds, Conference on Computer Vision and Pattern Recognition (CVPR) (2018)
  7652--7660.

\bibitem{Lang2019PointPillarsFE}
A.~H. Lang, S.~Vora, H.~Caesar, L.~Zhou, J.~Yang, O.~Beijbom, Pointpillars:
  Fast encoders for object detection from point clouds, in: Conference on
  Computer Vision and Pattern Recognition (CVPR), 2019, pp. 12689--12697.

\bibitem{Zhou2018VoxelNetEL}
Y.~Zhou, O.~Tuzel, Voxelnet: End-to-end learning for point cloud based 3d
  object detection, Conference on Computer Vision and Pattern Recognition
  (CVPR) (2018) 4490--4499.

\bibitem{dosovitskiy2017carla}
A.~Dosovitskiy, G.~Ros, F.~Codevilla, A.~Lopez, V.~Koltun, Carla: An open urban
  driving simulator, in: Conference on robot learning, PMLR, 2017, pp. 1--16.

\bibitem{xu2021opencda}
R.~Xu, Y.~Guo, X.~Han, X.~Xia, H.~Xiang, J.~Ma, Opencda: an open cooperative
  driving automation framework integrated with co-simulation, in: International
  Intelligent Transportation Systems Conference (ITSC), 2021, pp. 1155--1162.

\bibitem{vaswani2017attention}
A.~Vaswani, N.~Shazeer, N.~Parmar, J.~Uszkoreit, L.~Jones, A.~N. Gomez,
  {\L}.~Kaiser, I.~Polosukhin, Attention is all you need, Advances in neural
  information processing systems (NeurIPS) 30 (2017).

\bibitem{wang2020v2vnet}
T.-H. Wang, S.~Manivasagam, M.~Liang, B.~Yang, W.~Zeng, R.~Urtasun, V2vnet:
  Vehicle-to-vehicle communication for joint perception and prediction, in:
  European Conference on Computer Vision, Springer, 2020, pp. 605--621.

\bibitem{li2021learning}
Y.~Li, S.~Ren, P.~Wu, S.~Chen, C.~Feng, W.~Zhang, Learning distilled
  collaboration graph for multi-agent perception, Advances in Neural
  Information Processing Systems (NeurIPS) 34 (2021) 29541--29552.

\bibitem{xu2022v2x}
R.~Xu, H.~Xiang, Z.~Tu, X.~Xia, M.-H. Yang, J.~Ma, V2x-vit:
  Vehicle-to-everything cooperative perception with vision transformer, arXiv
  preprint arXiv:2203.10638 (2022).

\bibitem{Fcooper}
Q.~Chen, F-cooper: feature based cooperative perception for autonomous vehicle
  edge computing system using 3d point clouds, Proceedings of the 4th ACM/IEEE
  Symposium on Edge Computing (2019).

\bibitem{Kendall2017WhatUD}
A.~Kendall, Y.~Gal, What uncertainties do we need in bayesian deep learning for
  computer vision?, in: Advances in neural information processing systems
  (NeurIPS), 2017.

\bibitem{priornet_Malinin2018}
A.~Malinin, M.~J.~F. Gales, Predictive uncertainty estimation via prior
  networks, in: Advances in neural information processing systems (NeurIPS),
  2018.

\bibitem{bnn1_Mackay1991}
D.~J.~C. Mackay, A practical bayesian framework for backprop networks, Neural
  Computation (1991).

\bibitem{bnn2_Neal1995}
R.~M. Neal, Bayesian learning for neural networks, 1995.

\bibitem{Gal2016Unc}
Y.~Gal, Uncertainty in deep learning, in: PhD Thesis, 2016.

\bibitem{DeepEnsem}
B.~Lakshminarayanan, A.~Pritzel, C.~Blundell, Simple and scalable predictive
  uncertainty estimation using deep ensembles, in: Advances in neural
  information processing systems (NeurIPS), 2017.

\bibitem{DM_Feng2019onLidar}
D.~Feng, L.~Rosenbaum, F.~Timm, K.~C.~J. Dietmayer, Leveraging heteroscedastic
  aleatoric uncertainties for robust real-time lidar 3d object detection, IEEE
  Intelligent Vehicles Symposium (IV) (2019) 1280--1287.

\bibitem{DM_Meyer2019LaserNet}
G.~P. Meyer, A.~G. Laddha, E.~Kee, C.~Vallespi-Gonzalez, C.~K. Wellington,
  Lasernet: An efficient probabilistic 3d object detector for autonomous
  driving, in: Conference on Computer Vision and Pattern Recognition (CVPR),
  2019, pp. 12669--12678.

\bibitem{DM_Miller2019EvaluatingMS}
D.~Miller, F.~Dayoub, M.~Milford, N.~S{\"u}nderhauf, Evaluating merging
  strategies for sampling-based uncertainty techniques in object detection,
  International Conference on Robotics and Automation (ICRA) (2019) 2348--2354.

\bibitem{DM_Pan2020TowardsBP}
H.~Pan, Z.~Wang, W.~Zhan, M.~Tomizuka, Towards better performance and more
  explainable uncertainty for 3d object detection of autonomous vehicles,
  International Conference on Intelligent Transportation Systems (ITSC) (2020)
  1--7.

\bibitem{MD_Feng2020CamLidar}
D.~Feng, Y.~Cao, L.~Rosenbaum, F.~Timm, K.~C.~J. Dietmayer, Leveraging
  uncertainties for deep multi-modal object detection in autonomous driving,
  IEEE Intelligent Vehicles Symposium (IV) (2020) 877--884.

\bibitem{cls_unc2019TowardsBC}
V.~T. Vasudevan, A.~Sethy, A.~R. Ghias, Towards better confidence estimation
  for neural models, 2019, pp. 7335--7339.

\bibitem{Dempster2008AGO}
A.~P. Dempster, A generalization of bayesian inference, in: Classic Works of
  the Dempster-Shafer Theory of Belief Functions, 2008.

\bibitem{crf}
J.~D. Lafferty, A.~McCallum, F.~Pereira, Conditional random fields:
  Probabilistic models for segmenting and labeling sequence data, in:
  International Conference on Machine Learning, 2001.

\bibitem{ronneberger2015u}
O.~Ronneberger, P.~Fischer, T.~Brox, U-{N}et: Convolutional networks for
  biomedical image segmentation, in: International Conference on Medical image
  computing and computer-assisted intervention, Springer, 2015, pp. 234--241.

\bibitem{mink_choy20194d}
C.~Choy, J.~Gwak, S.~Savarese, 4d spatio-temporal convnets: Minkowski
  convolutional neural networks, in: Conference on Computer Vision and Pattern
  Recognition (CVPR), 2019, pp. 3075--3084.

\bibitem{ciassd}
W.~Zheng, W.~Tang, S.~Chen, L.~Jiang, C.-W. Fu, Cia-ssd: Confident iou-aware
  single-stage object detector from point cloud, in: AAAI, 2021.

\bibitem{xu2023opencda}
R.~Xu, H.~Xiang, X.~Han, X.~Xia, Z.~Meng, C.-J. Chen, J.~Ma, The opencda
  open-source ecosystem for cooperative driving automation research, arXiv
  preprint arXiv:2301.07325 (2023).

\bibitem{kingma2014adam}
D.~P. Kingma, J.~Ba, Adam: A method for stochastic optimization, in: ICLR,
  2014.

\bibitem{Wang2020InferringSU}
Z.~Wang, D.~Feng, Y.~Zhou, W.~Zhan, L.~Rosenbaum, F.~Timm, K.~C.~J. Dietmayer,
  M.~Tomizuka, Inferring spatial uncertainty in object detection, International
  Conference on Intelligent Robots and Systems (IROS) (2020) 5792--5799.

\bibitem{v2v_protocol}
F.~Arena, G.~Pau, An overview of vehicular communications, Future Internet 11
  (2019) 27.

\end{thebibliography}

\end{document}